\documentclass[journal]{IEEEtran}

\usepackage[T1]{fontenc}
\usepackage[utf8]{inputenc}
\usepackage{amsmath,amssymb,bm}
\usepackage{graphicx}
\usepackage{array}
\usepackage{multirow}
\usepackage{makecell}
\usepackage{booktabs}
\usepackage[table]{xcolor}
\usepackage{colortbl}
\usepackage{pifont}
\usepackage{algorithm}
\usepackage{algorithmic}
\usepackage{stfloats}
\usepackage[skins,breakable]{tcolorbox}
\usepackage{url}
\usepackage[colorlinks=true,linkcolor=black,citecolor=black,urlcolor=black]{hyperref}

\definecolor{shapecolor}{rgb}{0.0,0.5,0.0}
\definecolor{dkgreen}{rgb}{0,0.6,0}
\definecolor{mauve}{rgb}{0.58,0,0.82}
\begin{document}

\title{LookWise: Knowing When and Where to Look for Fine-Grained Visual Reasoning in Multimodal Large Language Models}

\author{Yuxiang~Shen, Hailong~Huang, Zhenkun~Gao, Xueheng~Li, Man~Zhou, Chengjun~Xie, Haoxuan~Che, Xuanhua~He, and Jie~Zhang%
\thanks{This work was supported by the Anhui Provincial Natural Science Foundation under Grant 2408085MD090.}%
\thanks{Yuxiang Shen, Hailong Huang, and Zhenkun Gao contributed equally to this work. Xuanhua He and Jie Zhang are the corresponding authors.}%
\thanks{Yuxiang Shen and Xueheng Li are with the Institute of Intelligent Machines, Hefei Institutes of Physical Science, Chinese Academy of Sciences, Hefei 230031, China, and also with the University of Science and Technology of China, Hefei 230026, China (e-mail: shenyuxiang\_xmu@163.com; lixueheng@mail.ustc.edu.cn).}%
\thanks{Man Zhou is with the University of Science and Technology of China, Hefei 230026, China (e-mail: manman@mail.ustc.edu.cn).}%
\thanks{Chengjun Xie and Jie Zhang are with the Institute of Intelligent Machines, Hefei Institutes of Physical Science, Chinese Academy of Sciences, Hefei 230031, China (e-mail: cjxie@iim.ac.cn; zhangjie@iim.ac.cn).}%
\thanks{Hailong Huang is with Zhejiang University, Hangzhou 310027, China (e-mail: huanghailong73@gmail.com).}%
\thanks{Zhenkun Gao is with East China Normal University, Shanghai 200241, China (e-mail: 51275901149@stu.ecnu.edu.cn).}%
\thanks{Haoxuan Che and Xuanhua He are with The Hong Kong University of Science and Technology, Hong Kong, China (e-mail: hche@ust.hk; xhecd@connect.ust.hk).}}

\markboth{IEEE Transactions on Circuits and Systems for Video Technology}
{Shen \MakeLowercase{\textit{et al.}}: LookWise: Knowing When and Where to Look}

\maketitle

\begin{abstract}
Multimodal Large Language Models (MLLMs)
are shifting towards "Thinking with Images"
by actively exploring image details. While effective, large-scale training is computationally expensive, which has spurred growing interest in lightweight, training-free solutions. However, existing training-free methods suffer from
two flaws: perceptual redundancy from indiscriminate cropping, which increases computational cost and introduces noise; and a drift between semantic intent and
spatial attention, which prevents accurate localization of user-focused regions. To address these challenges, we propose LookWise, a framework for adaptive visual reasoning. LookWise follows a two-stage pipeline: a confidence-based module decides when to look more carefully, and a semantic-guided localization module determines where to look. This design enables MLLMs to adaptively acquire fine-grained visual evidence without additional training. Experiments on fine-grained and high-resolution visual reasoning benchmarks show that LookWise consistently improves accuracy over strong baselines while achieving an approximately $4.0\times$ inference speedup over the search-based method ZoomEye, demonstrating robust cross-model generalization. Code is available at: \url{https://github.com/Xiaoxiang100/LookWise}.
\end{abstract}


\IEEEpeerreviewmaketitle

\section{Introduction}
\label{sec:intro}

In recent years, Multimodal Large Language Models (MLLMs) \cite{bai2025qwen2, wang2025internvl3, hurst2024gpt, comanici2025gemini} have made significant progress in visual-language understanding \cite{gao2026tpru, li2025llava, wang2026explore,tang2026video}. The prevailing paradigm, ``Thinking about Images,'' encodes images into static visual features and then applies language-based reasoning mechanisms such as Chain-of-Thought \cite{li2024surveying,wei2022chain}. 
However, this passive paradigm limits adaptive visual acquisition, as inputs are uniformly resized, cropped, or tiled into fixed visual tokens, producing static representations that cannot recover fine-grained details or adjust focus according to reasoning demands. As a result, performance often degrades on tasks involving tiny targets, subtle attributes, or high-resolution scenes \cite{li2025perception,wang2026divico}.

To address this limitation, a new paradigm termed ``Thinking with Images''~\cite{OpenAI2025Thinking} has emerged, empowering models with active perception capabilities. Instead of relying only on a single static encoding, models can dynamically decide when and where to zoom, crop, or scan images in response to the current query \cite{yu2025heurvidqa,zhao2026videoexpert}. This capability has been demonstrated by training-based approaches such as DeepEyes \cite{zheng2025deepeyes} and V-Star \cite{hosseini2024vstartrainingverifiersselftaught}. However, these methods require extensive computational resources and task-specific training data, motivating the exploration of training-free alternatives \cite{kim2025training, lee2025training} that enable adaptive visual reasoning at inference time.

Recent training-free methods attempt to acquire fine-grained visual details through inference-time strategies. Search-based methods, such as DC\textsuperscript{2}~\cite{wang2025divide} and ZoomEye~\cite{shen2025zoomeye}, iteratively evaluate image regions but incur 5--10$\times$ inference overhead. Single-pass methods, such as MLLMs-Know~\cite{zhang2025mllms}, rely on attention maps for efficient localization but often struggle with fine-grained and multi-object cases. These methods reveal two fundamental challenges: (1) determining \textit{when} visual enhancement is necessary, since indiscriminate cropping wastes computation and may introduce noise; and (2) identifying \textit{where} to precisely localize targets, especially when several objects compete for attention. Together, these questions of when to look and where to look constitute the core challenge of training-free adaptive visual reasoning.

To validate these hypotheses and understand the underlying 
mechanisms, we conducted controlled experiments across multiple benchmarks, revealing two critical patterns (as shown in Sec~\ref{sec:analysis}).

\noindent\textbf{Observation 1: Models know when to look, but current methods do not ask.} Existing methods usually crop all inputs under the assumption that finer visual details are always beneficial. However, our analysis shows that cropping can degrade performance in some cases, as it may fail to add useful information for simple queries and instead introduce interfering noise, as shown in \autoref{fig:phenomenon}. We further observe systematic differences in the model's confidence distributions across samples of varying difficulty, indicating that MLLMs inherently provide a signal for when more visual information is needed. This motivates our confidence-based strategy for deciding when to look more carefully.

\noindent\textbf{Observation 2: Attention-based methods struggle with where to look.} While attention maps provide an efficient way to locate visual regions, they often fail in multi-object scenarios. The model may attend to only one target, drift toward a salient distractor, or merge nearby instances into a single response. \autoref{fig:experiments} reveals that raw attention captures ``what is relevant'' rather than reliably resolving ``where each queried object is.'' This motivates our semantic-guided localization module, which decouples language intent from raw visual attention to determine where to look.

Building on these insights, we propose \textbf{LookWise}. Unlike prior methods that uniformly crop all inputs, LookWise dynamically determines whether cropping is beneficial and precisely localizes the queried targets.  Specifically, LookWise first decides when to look more carefully by using token-level confidence from the initial global prediction as an efficient indicator of information sufficiency. Once additional visual evidence is needed, it determines where to look through semantic-guided localization, which uses question-derived target semantics to guide visual attention toward the queried regions. In this way, LookWise achieves decision-aware zooming without iterative search or additional training.
\begin{figure}[H]
  \centering
  \includegraphics[height=0.47\textheight]{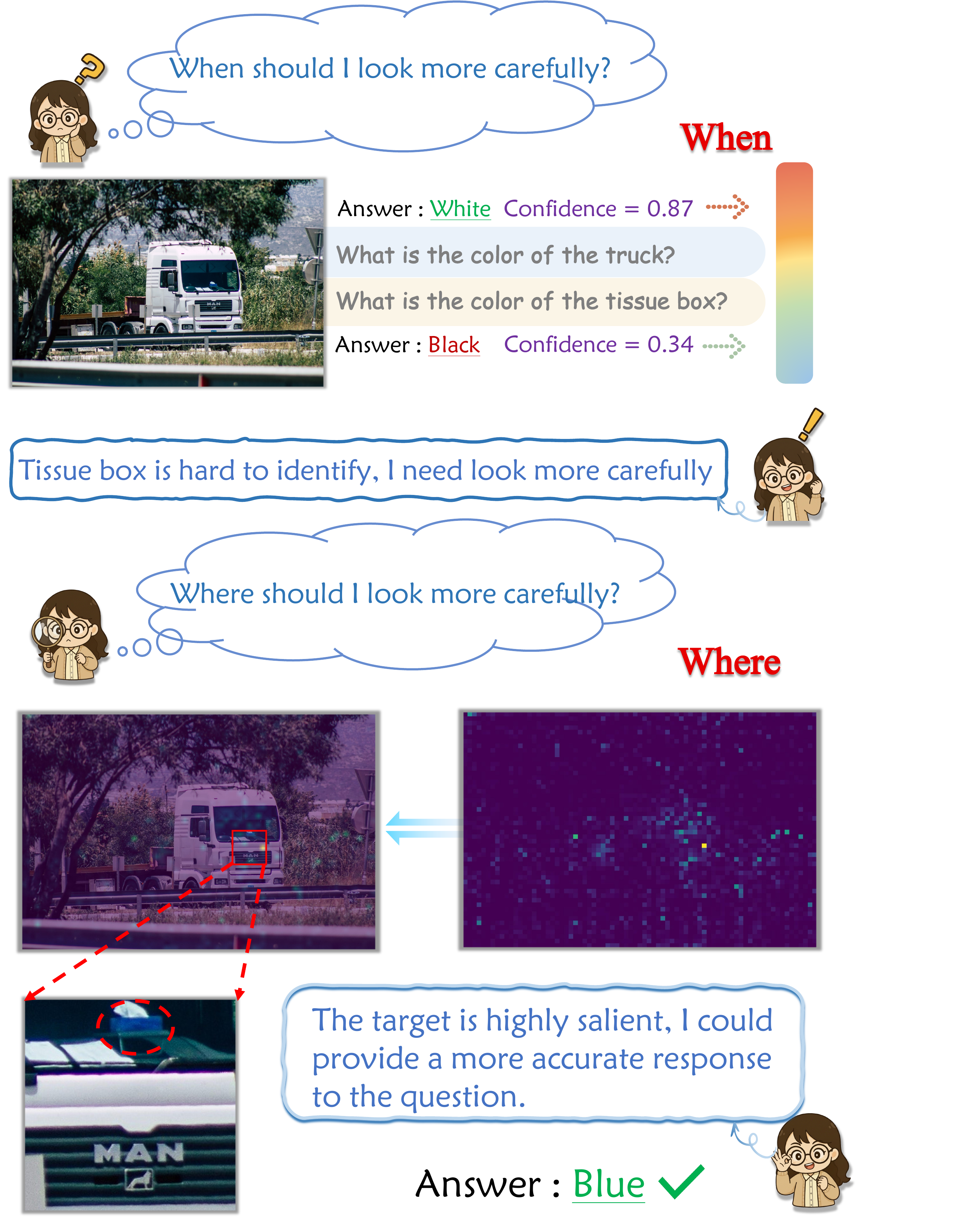}
  \caption{Illustration of LookWise: enabling MLLMs to know when and where to look for adaptive visual reasoning.}
  \label{fig:case}
\end{figure}

Our contributions can be summarized as follows:
(1) We identify two key deficiencies of existing training-free visual reasoning methods: indiscriminate cropping that causes perceptual redundancy, and raw attention localization that suffers from semantic drift in multi-object scenes. (2) We propose \textbf{LookWise}, a plug-and-play framework that combines confidence-based decision making with semantic-guided localization to answer when and where to look. (3) Extensive experiments demonstrate that LookWise achieves superior accuracy across multiple benchmarks while delivering a 4.0$\times$ speedup over the search-based method ZoomEye and generalizing across different MLLM backbones.

\section{Related Work}
\subsection{Multimodal Large Language Models}
Multimodal Large Language Models (MLLMs) have made significant progress by aligning visual perception with textual reasoning \cite{zhu2024multimodal}. In modern architectures, a pre-trained vision encoder is connected to a frozen large language model via learnable fusion modules. These modules range from simple linear projections \cite{liu2023visual, zhu2023minigpt} to query-based networks like Q-Formers \cite{dai2023instructblip}, which map visual features into the semantic space of the LLM.

However, traditional MLLMs typically encode images at a fixed, low resolution, leading to the loss of fine-grained details. This limits their performance on tasks involving small objects or complex scenes \cite{li2024monkey, wu2024v}. To preserve more visual information, recent models such as LLaVA-NeXT \cite{liu2024llavanext} and InternVL \cite{chen2024internvl} adopt dynamic high-resolution strategies. They divide a high-resolution image into smaller patches, encode each patch independently, and directly concatenate the resulting visual tokens.

While these high-resolution strategies improve image clarity, they still process image patches passively and uniformly. The language model must consume all generated visual tokens, regardless of whether a patch contains task-relevant evidence or background noise. This wastes computation and may still miss the critical details required by the query. Therefore, fine-grained MLLM inference requires an adaptive mechanism that knows when and where to look.

\subsection{Vision-Language Reasoning}
To overcome the limitations of fixed-resolution processing, recent research enables MLLMs to actively interact with visual content, similar to how human eyes focus on specific regions \cite{yu2025heurvidqa,zhao2026videoexpert}. Current approaches to this active perception generally fall into two categories: training-based and training-free methods. 

Training-based approaches, such as DeepEyes \cite{zheng2025deepeyes} and Pixel Reasoner \cite{wang2025pixel}, teach models how to crop and zoom through reinforcement learning or supervised step-by-step visual trajectories. While effective, these methods are highly expensive. They require massive amounts of specialized training data and computational resources. Moreover, the learned perception strategies are tightly bound to the specific model during training, making it difficult to apply them to other off-the-shelf MLLMs.

To avoid these high training costs, training-free methods attempt to simulate human-like zooming directly during the inference stage. Early training-free methods rely on iterative search. For example, ZoomEye \cite{shen2025zoomeye} crops images into smaller patches and explores them through a hierarchical tree. Although this search-based approach finds fine-grained details accurately, the multi-round evaluation process is extremely slow, resulting in high inference latency. More efficient approaches, such as ViCrop~\cite{zhang2025mllms}, use internal cross-attention maps to directly locate targets. However, they typically adopt an ``always-crop'' policy, focusing on \textit{where} to look while neglecting \textit{whether} zooming is needed. For visually simple tasks, this blind cropping wastes computation and often introduces background noise that misleads the model. Furthermore, they struggle to separate and locate targets accurately in complex, multi-object scenarios.

In contrast, LookWise unifies token confidence and attention maps for autonomous, decision-aware zooming. Token confidence determines whether more visual information is necessary before any extra visual operation is performed, while semantic-guided attention determines where to crop once visual enhancement is triggered. By deciding whether to crop before performing localization, LookWise retains the efficiency of attention-based methods and achieves competitive localization quality without the high latency of iterative search.

\subsection{Confidence Estimation and Adaptive Inference}
Confidence estimation allows models to evaluate their own uncertainty. In Large Language Models (LLMs), token-level probabilities serve as simple and effective indicators of how confident the model is about its predictions~\cite{fadeeva2024fact,zhang2025tokur}. When a model shows low confidence, it is more likely to hallucinate or make reasoning errors. Recent studies have introduced various methods to quantify this uncertainty, such as aggregating token probabilities~\cite{yaldiz2025not} or calculating entropy~\cite{xu2025tecp}, essentially helping models "know what they don't know."

Although confidence signals are widely used in text generation, they are rarely used to control visual computation in MLLMs. Many training-free methods still crop every input, wasting computation on visually simple samples. We use token confidence from the initial global prediction as a routing signal. If the confidence is high, the model answers directly; otherwise, LookWise activates semantic-guided localization and performs a second prediction with local visual details. This adaptive strategy reduces redundant visual processing and improves inference efficiency.
\section{Methods}
\subsection{Analysis: When and Where to Look}
\label{sec:analysis}
Before introducing LookWise, we analyze the fundamental limitations of existing training-free methods in determining when and where to crop. Our analysis aims to answer two critical questions: (1) Can MLLMs know when additional visual information is necessary? (2) What causes attention-based localization to fail on fine-grained tasks?

\subsubsection{Can MLLMs know when to look?}
\begin{figure*}[!t]
    \centering
    \includegraphics[width=0.98\textwidth]{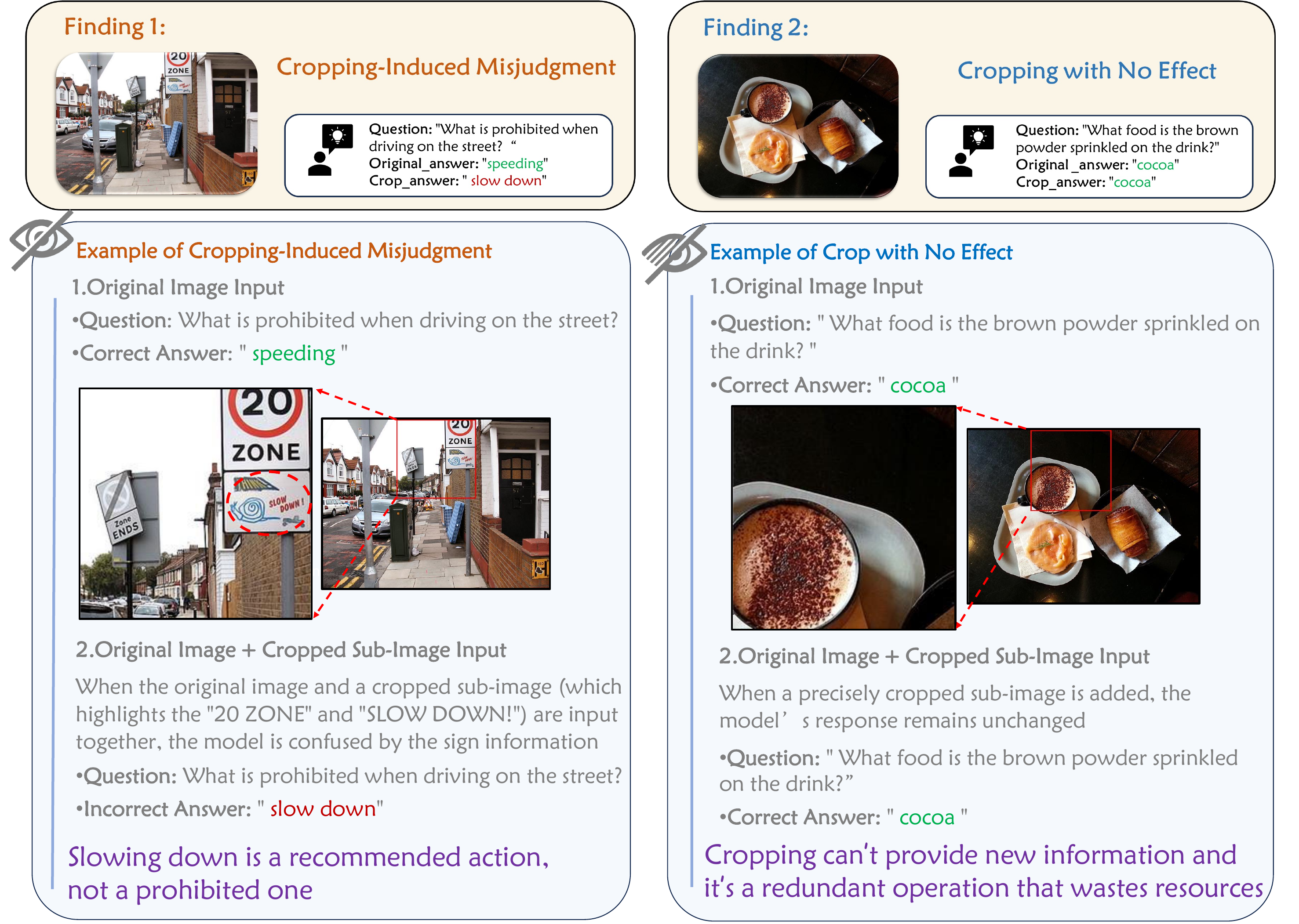}
    \caption{Limitations of indiscriminate cropping. 
Left: Cropping a region containing the ``SLOW DOWN'' sign misleads the model to answer ``slow down,'' even though the question asks what action is prohibited and the correct answer is ``speeding.'' 
Right: Cropping the drink area adds no new visual information, but the model's prediction remains unchanged while additional computation is introduced.}
    \label{fig:phenomenon}

\end{figure*}

In the \textbf{Thinking with Images} paradigm, some training-based methods learn when to look through additional supervision or reinforcement learning. Most training-free methods, however, adopt indiscriminate cropping and re-encoding for all samples, neglecting the critical question of when to crop or zoom. For visually simple samples, forced cropping incurs unnecessary computational overhead and may introduce redundant or noisy local information that misleads the model into incorrect answers. \autoref{fig:phenomenon} presents two representative examples. First, a crop containing a salient ``SLOW DOWN'' sign misleads the model to answer with the sign text, overriding the correct global prediction. Second, when the target ``cocoa'' is already clear in the global image, adding a crop provides little new information but still increases inference cost.

\begin{figure}[H]
  \includegraphics[width=0.95\columnwidth]{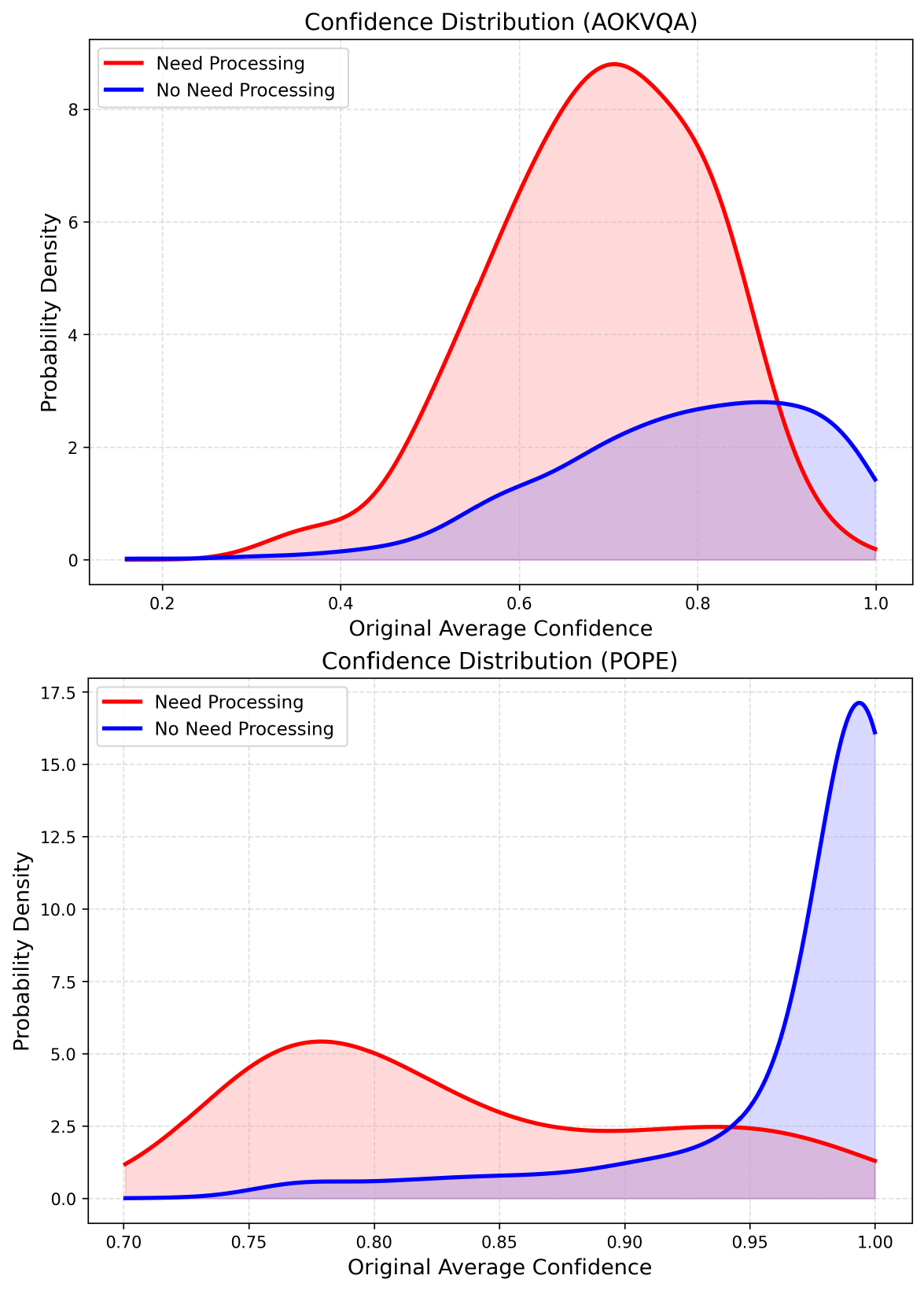}
\caption{Probability density distributions of initial average confidence on the AOKVQA and POPE datasets.}
  \label{fig:confidence}
\end{figure}

Additional visual evidence is often expected to improve the prediction confidence of MLLMs. However, this assumption has not been systematically validated across different samples. To investigate this, we conduct a token-level confidence analysis before and after cropping. Based on the confidence change after cropping, we categorize samples into two groups:
\emph{(i) Need Processing}, where confidence increases ($Score_{\mathrm{crop}} > Score_{\mathrm{org}}$), and
\emph{(ii) No Need Processing}, where confidence remains unchanged or decreases ($Score_{\mathrm{crop}} \leq Score_{\mathrm{org}}$).
The confidence distributions of the two groups are shown in \autoref{fig:confidence}.

As shown in \autoref{fig:confidence}, the \emph{Need Processing} and \emph{No Need Processing} samples exhibit a clear separation in their original average confidence distributions. 
On both AOKVQA and POPE, the \emph{No Need Processing} samples show higher initial confidence, indicating that the original image provides sufficient visual evidence for prediction. 
In contrast, the \emph{Need Processing} samples are concentrated in lower-confidence regions, suggesting that additional visual details are more beneficial when the model's initial prediction is uncertain.

This analysis reveals that cropping has a heterogeneous effect on model confidence: while some samples benefit from additional details, many show unchanged or even reduced confidence. Importantly, the initial confidence reflects the model's need for additional visual information, enabling a simple and reliable criterion for deciding \textbf{when to look}.

\subsubsection{Why does attention fail to decide where to look?} 
MLLMs possess inherent visual grounding capabilities \cite{zhang2025mllms}, and their spatial attention enables more efficient localization than search-based methods such as DC$^2$ \cite{wang2025divide} and ZoomEye~\cite{shen2025zoomeye}. However, directly using raw attention maps introduces two challenges for precise target localization.

The first problem occurs when a question involves multiple objects, e.g., ``Is the bicycle on the left or right side of the motorcycle?'' The attention response may spread across several entities and fail to isolate the queried object. We address this by first extracting the main target from the question using few-shot in-context learning (ICL). The text tokens of this target are then used as queries over image tokens, producing a semantic-guided attention map that focuses on the object of interest more precisely.

The second problem appears when many objects of the same category are close together, e.g., ``How many people?'' The attention map often merges nearby instances into one region, making instance-level reasoning difficult. We mitigate this issue by applying attention thresholding to generate candidate boxes, followed by an NMS-inspired overlap filtering step to remove redundant regions. The model then produces the final answer using the refined visual evidence. 

By addressing these two issues, LookWise obtains more reliable target regions and provides a practical answer to where to look in complex scenes.

\subsection{LookWise: Overview}
\begin{figure*}[!t]
    \centering
    \includegraphics[width=0.98\textwidth]{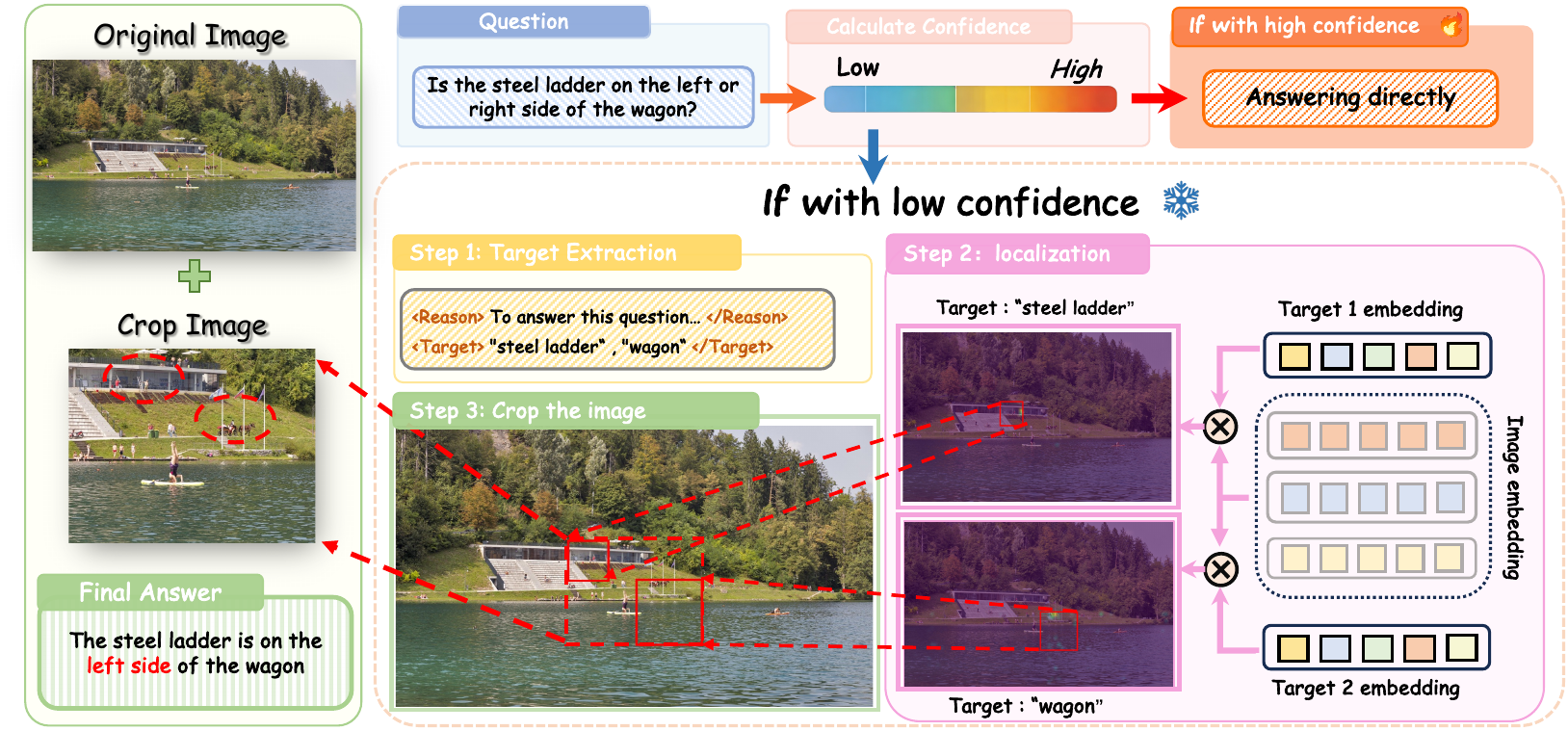}
\caption{
Overview of the LookWise framework. The model evaluates token confidence to decide when to look more carefully. If uncertain, it extracts key textual targets and uses them as queries in cross-attention over visual tokens to determine where to crop. The selected local regions are then provided together with the global image to produce the final answer.
}
    \label{fig:framework}

\end{figure*}

To address the limitations of existing training-free methods, we propose \textbf{LookWise}, a two-stage framework for adaptive visual reasoning. The first stage is a \textit{confidence-based module} that decides when to crop, avoiding redundant processing for simple instances. The second stage is a \textit{semantic-guided localization module} that determines where to crop by integrating the target semantics extracted from the question with the model's spatial attention. Together, these two stages enable MLLMs to look more carefully only when needed and at the regions most relevant to the query. The overall workflow of LookWise is illustrated in \autoref{fig:framework}.

\subsection{Confidence-Based Module: Determining When to Look}
\label{sec:when_to_fuse}
For a given input image $I$ and query $Q$, the model first performs an initial inference to obtain a preliminary answer sequence, denoted as $Answer_{\mathit{pre}}$. Relying solely on the global view may be insufficient for fine-grained reasoning, but cropping every sample is inefficient. We therefore design a confidence-based module to determine whether a second visual enhancement stage should be triggered. Unlike explicit prompting methods that require an additional generation step to make a routing decision, our strategy directly uses the token probabilities from the initial forward pass. This allows the model to estimate its own uncertainty with virtually zero extra computational cost.

\subsubsection{Model Confidence Quantification.} We quantify the confidence of the model by utilizing the average confidence of the tokens in the generated answer sequence. Specifically, the sequence is defined as
\begin{equation}
    Y = \{y_1, y_2, \dots, y_T\},
\end{equation}
where $y_t$ represents the $t$-th token, and $P(y_t|\cdot)$ denotes the probability of the model generating this token at step $t$. We define the confidence score $C$ as the arithmetic mean of the Softmax probabilities of all tokens in the sequence. This confidence score $C$ intuitively reflects the model's epistemic uncertainty regarding the preliminary answer $Answer_{\mathit{pre}}$.
\begin{equation}
    C = \frac{1}{T} \sum_{t=1}^{T} P(y_t|\cdot).
\end{equation}

\subsubsection{Optimal Threshold Selection.}
A simple yet effective approach is to use a fixed decision threshold $\tau$. As demonstrated in Section~\ref{sec:Confidence Sensitivity Analysis}, cropping only samples with confidence scores below this threshold improves performance while reducing computational overhead.

When representative validation data are available and closely aligned with the test distribution, we further consider a principled threshold optimization strategy. Specifically, threshold selection is formulated as a statistical decision problem, where the optimal $\tau$ is obtained by maximizing the following utility function:
\begin{equation}
    \label{eq:utility_function}
    J_{\lambda}(\tau) = \text{TPR}(\tau) - \lambda \cdot \text{FPR}(\tau), \quad \lambda > 0.
\end{equation}
Here, TPR and FPR denote the true positive rate and false positive rate, respectively, and $\lambda$ serves as a cost-weighting hyperparameter that balances potential performance gains against unnecessary local inspection.

This optimization strategy assumes access to representative validation data. \textbf{In open-domain settings where such data cannot be reliably obtained, employing a high fixed threshold remains a stable and effective alternative,} as evidenced by the results in ~\autoref{tab:conf_sensitivity}. These observations further demonstrate the robustness and generalization capability of our approach.

\subsubsection{Final Decision Process.}
Based on the optimal decision boundary, we perform the following binary decision:
(1) $C \ge \tau$: the model exhibits high confidence on the global view.
In this case, we consider the available visual information sufficient for reasoning and directly output $Answer_{\mathit{pre}}$ as the final answer.
(2) $C < \tau$: the model shows high uncertainty, indicating that the current visual information may be insufficient to support fine-grained reasoning.
We therefore trigger the subsequent semantic-guided localization module to acquire high-resolution local evidence.

\subsection{Semantic-Guided Localization Module: Determining Where to Look}
Once cropping is triggered, the model must precisely localize the target region. However, relying exclusively on raw visual attention often leads to attention misalignment. In complex scenes or multi-object queries, unguided attention may activate irrelevant backgrounds or conflate distinct instances. 

To address this limitation, we introduce a semantic-guided localization module that uses explicit language intent to guide spatial localization. The process has two steps. First, we decouple the core physical target from the user query. Second, the text tokens of the extracted target act as queries over image tokens, producing a semantic-guided attention map. This text-to-vision guidance translates the abstract query intent into a concrete bounding box for local evidence extraction.
\subsubsection{Semantic Decoupling (What to Localize).} 
To accurately extract key visual entities, We design a prompt template based on Chain-of-Thought. We guide the model to first generate a \texttt{<Reason>} tag before outputting the final \texttt{<target>}.

Formally, given a question $Q$, the model outputs a semantic target set $E = \{e_1, e_2, \dots\}$. For example, for "What kind of animal is on the red sign?", the model generates: \texttt{<Reason>The question asks about something on a sign. The core subject to locate is the sign.</Reason><Target>red sign</Target>}. Then it extracts $E = \{\text{"red sign"}\}$.

To illustrate this extraction capability across complex scenarios, we integrate a standardized prompt template and provide comprehensive few-shot examples, as shown below.

\begin{tcolorbox}[
    title=Target Extraction Prompt,
    width=\linewidth,
    breakable,
    enhanced,
    fonttitle=\bfseries,
    colback=white,
    colframe=black!70,
    boxrule=0.8pt
]
\footnotesize
\textbf{Question Template:} \\
Based on the question, identify ONLY the primary, physical objects or subjects mentioned. Do not include adjectives, locations, or states.

Please respond using \texttt{<Reason>} and \texttt{<target>} tags.

\textbf{Guidelines:}
\begin{itemize}
    \item A 'target' must be a simple, concrete noun (e.g., 'car', 'person', 'table').
    \item For multiple distinct subjects, separate them with a comma (e.g., 'cat, dog').
\end{itemize}

\textbf{Question:} What is the brand name on the laptop? \\
\textbf{Response:} \texttt{<Reason>}The question asks about a brand name found on a laptop. The core physical subject is the laptop.\texttt{</Reason>}\texttt{<target>}laptop\texttt{</target>}

\textbf{Question:} How many people are wearing hats in the image? \\
\textbf{Response:} \texttt{<Reason>}The question asks to count people who are wearing hats. The primary subjects are people and hats.\texttt{</Reason>}\texttt{<target>}people, hats\texttt{</target>}

\textbf{Question:} Is the cat on the left or right side of the wooden chair? \\
\textbf{Response:} \texttt{<Reason>}The question asks about the position of a cat relative to a wooden chair. The core physical subjects are the cat and the chair.\texttt{</Reason>}\texttt{<target>}cat, wooden chair\texttt{</target>}

\textbf{Question:} What kind of food is on the white plate? \\
\textbf{Response:} \texttt{<Reason>}The question asks about the type of food on a plate. The primary physical subjects are the food and the plate.\texttt{</Reason>}\texttt{<target>}food, white plate\texttt{</target>}

\textbf{Question:} What is the man in the blue shirt holding in his hand? \\
\textbf{Response:} \texttt{<Reason>}The question asks about an object held by a man. The core physical subjects are the man and the object he is holding.\texttt{</Reason>}\texttt{<target>}man, object\texttt{</target>}

\textbf{Question:} What time is displayed on the clock on the wall? \\
\textbf{Response:} \texttt{<Reason>}The question asks for the time displayed on a clock. The core physical subject is the clock.\texttt{</Reason>}\texttt{<target>}clock\texttt{</target>}
\end{tcolorbox}

\subsubsection{Spatial Attention Mapping (Where to Localize).} 
After identifying the semantic target $E$, we instantiate a target-specific localization prompt as ``where is the \texttt{<target>}''. We then feed this prompt to the MLLM and extract the cross-attention responses between the target tokens and image tokens to obtain a target-conditioned attention map. Compared with attention derived from the original question, this target-specific formulation suppresses irrelevant objects and reduces attention drift in multi-object scenes.

\noindent\textbf{2D Attention Map Generation.}
Let $Z_{\text{img}}$ denote the visual token sequence obtained from the image patches, and let $Z_{\text{text}}^{E}$ denote the text tokens corresponding to the extracted target. We extract the cross-attention responses from the MLLM and average them across all attention heads:
\begin{equation}
A_{\text{1D}} = \operatorname{Agg}\!\left( \operatorname{Attn}(Z_{\text{text}}^{E} \rightarrow Z_{\text{img}}) \right)
\label{eq:attention_map}
\end{equation}

The resulting vector $A_{\text{1D}}$ corresponds to the flattened visual token grid. We reshape it according to the patch layout of the vision encoder to obtain a 2D attention map $A_{\text{map}} \in \mathbb{R}^{h \times w}$, where each value indicates the attention intensity of the corresponding image region.

\noindent\textbf{Adaptive Multi-Scale Sliding Window.}
Given the spatial attention map $A_{\text{map}}$, our goal is to translate this activation grid into a precise bounding box. To accommodate the variance in input resolutions, we dynamically set an adaptive base dimension $S_{\text{base}}$ (typically $224 \times 224$, scaling to $448 \times 448$ for ultra-high-resolution images) based on the original image size. 

Building upon this adaptive base, we apply a predefined set of scaling ratios $R = \{1.0$, $1.2$, $1.4$, $1.6$, $1.8$, $2.0$, $4.0$, $6.0\}$. For each ratio $r \in R$, the candidate extraction size is calculated as $S_r = S_{\text{base}} \times r$. This physical dimension is then projected onto the $A_{\text{map}}$ grid to determine a corresponding window covering $w_r \times h_r$ attention blocks. We slide this window across the 2D grid to find the position $(x^*, y^*)$ that yields the maximum attention sum $V_{\max}$. 

To select the most fitting scale, we evaluate the \textit{localization sharpness} (or contrast) $\Delta_r$ for the peak window of each ratio. The sharpness is calculated by taking the difference between the peak attention $V_{\max}$ and the average attention of its four adjacent surrounding windows $V_{\text{adj}}$, normalized by the window area:
\begin{equation}
\Delta_r = \frac{V_{\max} - \operatorname{Mean}(V_{\text{adj}})}{w_r \times h_r}
\end{equation}
Intuitively, a higher $\Delta_r$ means the window tightly encapsulates the highly activated target while excluding the unactivated background. We select the optimal ratio $r^*$ that maximizes this sharpness metric.

\noindent\textbf{Coordinate Mapping.}
Once the optimal grid window is found, we calculate its geometric center and linearly map it back to the original high-resolution image space. Based on the selected box size, we deduce the final pixel coordinates $(x_1, y_1, x_2, y_2)$ and clamp them to the image boundaries. The original image $I$ is then cropped accordingly to produce a detail-preserving local view $I_{\text{crop}}$ for the final prediction stage.

\subsubsection{Dealing with Multi-instances.} 
For queries involving multiple instances of the same category (e.g., "How many people?"), we introduce an attention-based post-processing scheme to separate closely spaced targets. 
Specifically, an attention score threshold is applied to filter foreground regions, with high-response areas identified as candidate boxes. To remove highly overlapping boxes, we apply an NMS-inspired \cite{neubeck2006efficient} deduplication step: boxes with an IoU greater than 0.5 are treated as duplicates and pruned.  
This process effectively reduces spatial redundancy, enabling accurate localization and counting of individual objects.

\section{Experiment}
\subsection{Implementation Details}
\noindent\textbf{Experimental Setup \& Network Configuration.}
To validate the effectiveness of our approach, we integrate the LookWise framework with two widely-used open-source models: Qwen2.5VL-3B \cite{bai2025qwen2} and LLaVA-1.5-7B \cite{liu2023visual}. In our experiments, we set the maximum input resolution for Qwen to 3,211,264 pixels. All experiments are conducted on NVIDIA H200 GPUs. 
For attention map acquisition, we extract attention scores from specific layers to isolate visual regions relevant to the target entity. Specifically, for Qwen2.5-VL, we extract attention scores from Layer 22, utilizing the last token of the target sequence as the query. For LLaVA-1.5, we use attention weights from Layer 14, averaging the attention scores of all tokens comprising the final word of the target phrase to create a stable spatial representation. 
We conduct a comprehensive comparison against both the original baselines and several state-of-the-art methods, including training-free approaches (e.g., ZoomEye \cite{shen2025zoomeye}, MLLMs-Know \cite{zhang2025mllms}, DC$^2$ \cite{wang2025divide}, and ViCrop \cite{zhang2023towards}) and training-based models (e.g., PixelReasoner \cite{wang2025pixel}).

\noindent\textbf{Datasets and Benchmarks.}
We evaluate our model across various dimensions using four benchmarks. First, we select \textbf{AOKVQA} \cite{schwenk2022okvqa} and \textbf{POPE} \cite{li2023evaluating} to assess general reasoning capabilities on knowledge-intensive tasks. Furthermore, to validate the effectiveness of our confidence-based decision module on fine-grained features, we introduce two challenging high-resolution datasets: \textbf{V*-Bench} \cite{wu2024v} and \textbf{HR-Bench} \cite{wang2025divide}. These benchmarks emphasize fine-grained attribute recognition and complex spatial reasoning, strictly evaluating the model's ability to localize tiny visual details.

\subsection{Qualitative Analysis of Semantic-Guided Localization}
To visually demonstrate the effectiveness of the proposed where-to-look mechanism, we compare the attention maps and final bounding regions generated by LookWise with those from a baseline attention-based localization method (MLLMs-Know). 
\begin{figure}[!t]
  \includegraphics[width=\columnwidth]{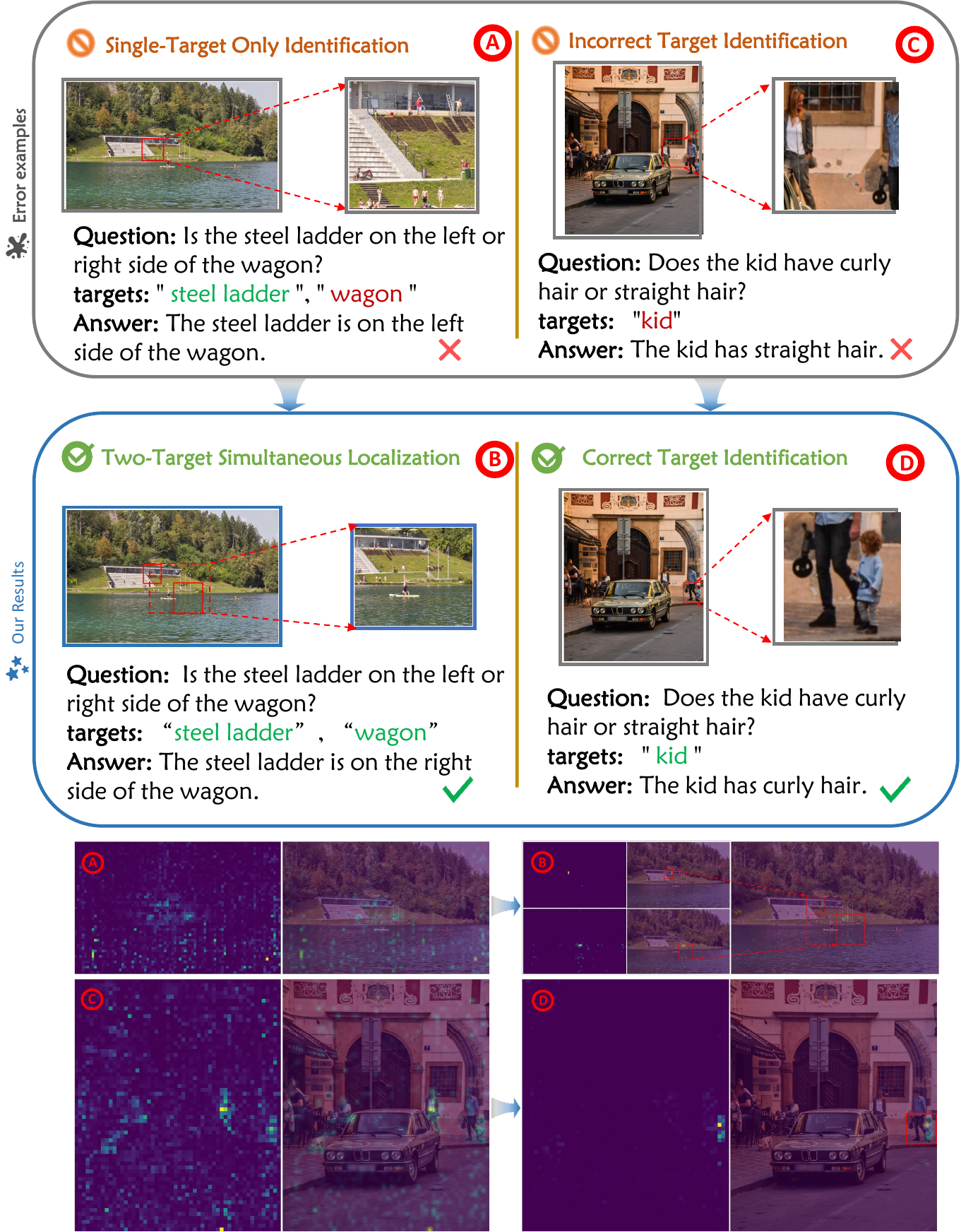}
\caption{Comparison of localization precision. The baseline model often suffers from single-target bias (A) and attention drift (C). In contrast, LookWise explicitly decouples multiple queried targets to ensure complete coverage (B) and accurately identifies the intended subject (D). The attention maps show that semantic-guided localization suppresses noisy activations and maintains precise spatial focus.}
  \label{fig:experiments}
\end{figure}
\subsubsection{Accurate multi-object localization:}In~\autoref{fig:experiments}(A), the baseline suffers from a single-object bias, attending only to the `steel ladder' while missing the `wagon'. In contrast, LookWise explicitly decouples multiple targets, activating both the `steel ladder' and `wagon'. As shown in ~\autoref{fig:experiments}(B), the resulting union of bounding boxes effectively covers both objects, providing the MLLM with a complete and precise local visual source to accurately reason about their spatial relationship.

\subsubsection{Robust target identification:}
In ~\autoref{fig:experiments}(C), the baseline's attention drifts to the dominant adult in the scene (incorrect target). Guided by our prompt-driven target extraction, LookWise accurately focuses on the "child", ensuring the extracted region aligns with the intended semantic target, as shown in~\autoref{fig:experiments}(D).
These qualitative results demonstrate that semantic-guided localization significantly mitigates attention diffusion and drift, ensuring that the visual enhancement process relies on accurate local evidence.

\subsection{Quantitative Analysis}
Our method shows strong performance on benchmarks that require fine-grained visual evidence, such as HR-Bench and V*-Bench. Although LLaVA operates at a fixed resolution, which naturally restricts its attention granularity, LookWise still brings notable improvements: \textbf{+11.25\%} on HR-Bench 4K, \textbf{+9.87\%} on HR-Bench 8K, and \textbf{+14.12\%} on V*-Bench. When integrated with Qwen2.5-VL, LookWise reaches a state-of-the-art \textbf{70.00\%} on HR-Bench 8K, exceeding the baseline by \textbf{11.12\%} and outperforming ZoomEye (65.63\%). These gains indicate that adaptive visual reasoning can compensate for resolution bottlenecks while avoiding unnecessary computation on irrelevant regions.

For reasoning-heavy benchmarks such as AOKVQA and POPE, the improvements are more moderate, suggesting that their main bottleneck lies in logical reasoning and commonsense knowledge rather than visual resolution. Nevertheless, the consistent gain, e.g., improving Qwen2.5-VL to \textbf{73.10\%} on AOKVQA, shows the benefit of confidence-based routing over indiscriminate cropping. By looking more carefully only on uncertain samples, LookWise preserves the global context for easy cases and provides additional details for visually ambiguous ones. Overall, the results show that the proposed inference strategy improves fine-grained perception while maintaining robustness on general visual reasoning tasks.
\begin{table*}[!t]
\centering
\small
\renewcommand{\arraystretch}{1.2}
\setlength{\tabcolsep}{3pt}

\definecolor{bluebest}{rgb}{0.65, 0.81, 0.95}
\definecolor{bluesecond}{rgb}{0.88, 0.95, 1.0}
\definecolor{redbest}{rgb}{0.953, 0.671, 0.639}
\definecolor{redsecond}{rgb}{0.984, 0.886, 0.875}
\definecolor{graybg}{gray}{0.92}

\newcommand{\bbest}{\cellcolor{bluebest}}
\newcommand{\bsec}{\cellcolor{bluesecond}}
\newcommand{\rbest}{\cellcolor{redbest}}
\newcommand{\rsec}{\cellcolor{redsecond}}

\providecommand{\improvement}[1]{\textbf{\textcolor{red}{+#1}}}

\newcolumntype{Y}{>{\centering\arraybackslash}p{1.8cm}}
\newcolumntype{Z}{>{\centering\arraybackslash}p{2.1cm}}
\caption{
\textbf{Comparison with state-of-the-art methods.}
Highest results are marked in darker colors (\colorbox{bluebest}{Blue} for LLaVA, \colorbox{redbest}{Red} for Qwen).
The \colorbox{graybg}{gray rows} show the absolute improvement ($\Delta$) of \textbf{Ours} over the \textbf{Baseline}.
$^{*}$ Results are reproduced by us using official implementations.
}
\label{tab:final_red_std_gain}
\resizebox{\linewidth}{!}{
    \begin{tabular}{l l Y Z Z Z Z Z}
    \toprule
    \multicolumn{1}{c}{\textbf{Model}} & \multicolumn{1}{c}{\textbf{Method}} & \textbf{Training-free} 
    & \textbf{AOKVQA} & \textbf{POPE} & \textbf{V$^*$ Bench} 
    & \textbf{HR-Bench 4K} & \textbf{HR-Bench 8K} \\
    \midrule

    \multirow{7}{*}{LLaVA-v1.5-7B}
     & Baseline$^{*}$~\cite{liu2023visual} & \ding{51}
       & 71.00 & 86.98 & 48.68 & 36.13 & 32.13 \\
     & DC$^2$~\cite{wang2025divide} & \ding{51}
       & - & - & 57.60 & - & 39.50 \\
     & VisCrop~\cite{zhang2023towards} & \ding{51}
       & - & - & 62.30 & 46.25 & 35.75 \\
     & MLLMs-Know$^{*}$~\cite{zhang2025mllms} & \ding{51}
       & \bsec 72.31 
       & 87.25 
       & 56.02 & 44.38 & 37.25 \\
     & ZoomEye~\cite{shen2025zoomeye} & \ding{51}
       & 70.56 
       & \bbest 88.94 
       & \bbest 83.25 
       & \bbest 49.88 
       & \bbest 48.63 
       \\
     & \textbf{Ours (LookWise)} & \ding{51}
       & \bbest 72.90 
       & \bsec 87.37  
       & \bsec 62.80  
       & \bsec 47.38  
       & \bsec 42.00  
       \\
       \rowcolor{graybg}
       & \quad \textbf{$\Delta$ vs. Baseline} & 
       & \improvement{1.90}  
       & \improvement{0.39}  
       & \improvement{14.12} 
       & \improvement{11.25} 
       & \improvement{9.87}  
       \\
    \midrule

    \multirow{6}{*}{Qwen2.5VL-3B}
     & Baseline$^{*}$~\cite{bai2025qwen2} & \ding{51}
       & 71.44 & 87.20 & 75.90 & 67.50 & 58.88 \\
     & Pixel Reasoner~\cite{wang2025pixel} & \ding{55}
       & - & - & 84.82 & - & 66.00 \\
     & MLLMs-Know$^{*}$~\cite{zhang2025mllms} & \ding{51}
       & \rsec 71.62 
       & \rbest 89.12 
       & 75.90 & 66.36 & 64.88 \\
     & ZoomEye~\cite{shen2025zoomeye} & \ding{51}
       & 71.26 
       & \rsec 88.93 
       & \rbest 89.01 
       & \rsec 70.13 
       & \rsec 68.38 
       \\
     & \textbf{Ours (LookWise)} & \ding{51}
       & \rbest 73.10 
       & \rbest 89.12 
       & \rsec 86.38  
       & \rbest 73.25 
       & \rbest 70.00 
       \\
       \rowcolor{graybg}
       & \quad \textbf{$\Delta$ vs. Baseline} & 
       & \improvement{1.66}  
       & \improvement{1.92}  
       & \improvement{10.48}  
       & \improvement{5.75}  
       & \improvement{11.12} 
       \\
    \bottomrule
    \end{tabular}
}

\end{table*}

\subsection{Efficiency analysis}
\subsubsection{Overall Inference Efficiency}
\begin{figure}[!t]
  \centering
  \includegraphics[width=1\columnwidth]{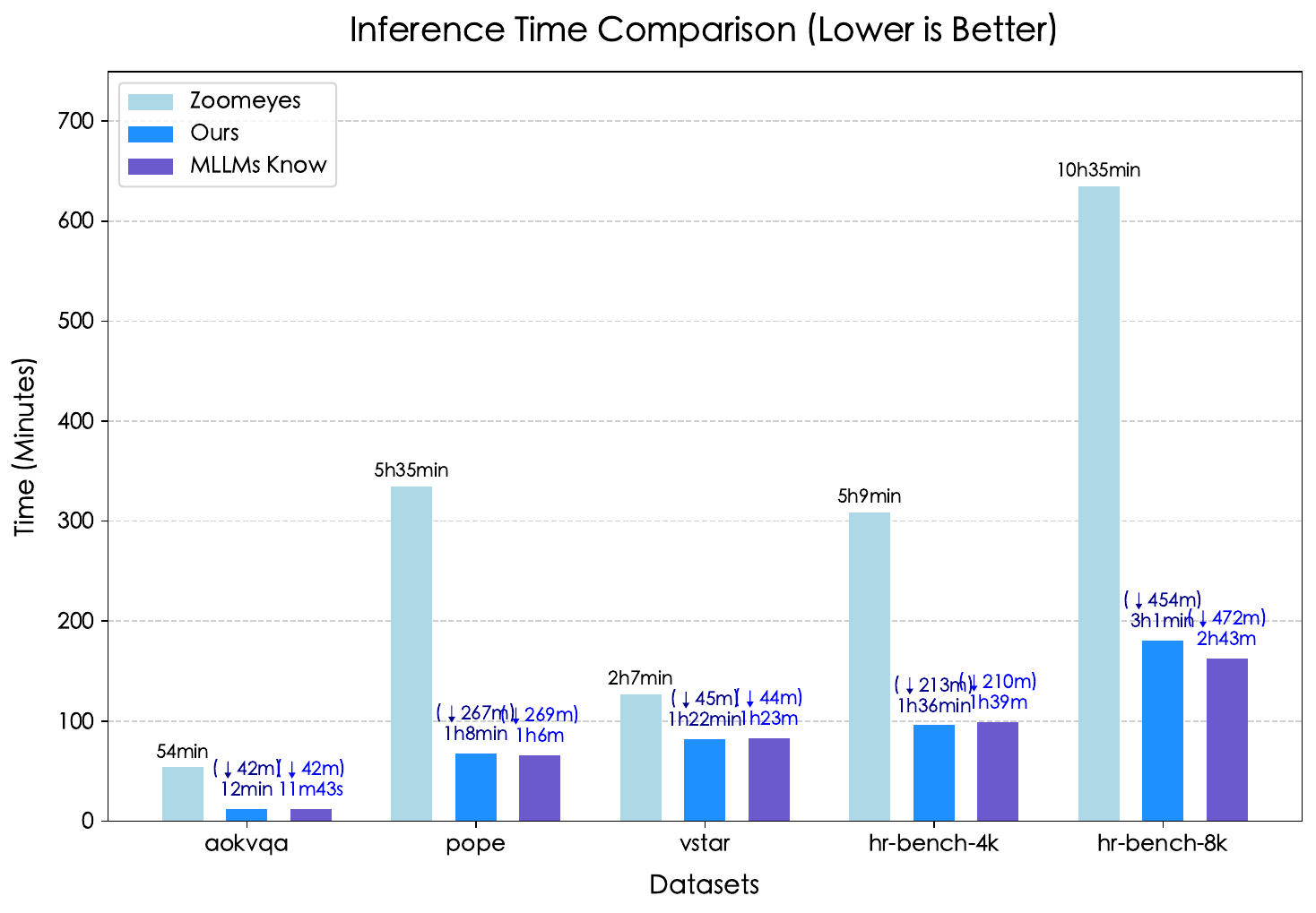}
  \caption{Comparison of total GPU inference time (in minutes) across different methods (measured on NVIDIA H200 GPUs).}
  \label{fig:time}
\end{figure}

Inference efficiency is an important factor for training-free methods. We evaluate LookWise against two representative baselines: \textbf{ZoomEye} (the current SOTA) and \textbf{MLLMs-Know} (an attention-based method).
ZoomEye organizes the image into a hierarchical tree and performs iterative search to simulate zooming. Although effective, this global traversal becomes increasingly expensive. 
As shown in \autoref{fig:time}, processing HR-Bench 8K requires more than 10 hours.
In contrast, LookWise localizes informative regions directly through semantic-guided attention, avoiding exhaustive search over the full image. On HR-Bench 8K, this yields an approximately $4.0\times$ speedup compared with ZoomEye, while maintaining competitive performance.

Furthermore, LookWise runs at a similar speed to MLLMs-Know across all evaluated datasets, while achieving higher accuracy on challenging high-resolution benchmarks. For example, LookWise reaches 86.38\% on V*-Bench and 70.00\% on HR-Bench 8K, compared with 75.90\% and 64.88\% achieved by MLLMs-Know, respectively. These results suggest that the improvement mainly comes from more effective allocation of local visual computation rather than increased inference cost.

\subsubsection{Contrast with LLM-Prompting Strategy} 

An intuitive alternative to our decision strategy is to directly prompt the MLLM to make the choice, e.g., asking "\textit{Do you need to look more carefully to answer this question? Answer Yes or No}". However, we demonstrate that utilizing token confidence is significantly more efficient than direct prompting.

To clearly illustrate the efficiency gap, we compare the inference pipelines of both methods in Algorithm \ref{alg:workflow}.

\begin{algorithm}
\caption{Inference Workflow: LLM-Prompting vs. Ours}
\label{alg:workflow}
\begin{algorithmic}[1]
\STATE \textcolor{gray}{\textit{\# Method 1: LLM-Prompting}}
\STATE $Decision \leftarrow$ \textbf{LLM\_Generate}($I_{global}$, \textit{"Need local details?"})
\IF{$Decision == \text{"No"}$}
    \STATE \textcolor{blue}{\textit{\# Still requires 1 round to answer}}
    \STATE \textbf{Return} \textbf{LLM\_Generate}($I_{global}$, $Question$)
\ELSE
    \STATE \textcolor{blue}{\textit{\# Decision is "Yes": Extract local details via attention}}\STATE $I_{local} \leftarrow$ \textbf{Target\_Guided\_Extract}($I_{global}$)
    \STATE \textbf{Return} \textbf{LLM\_Generate}($I_{global}$, $I_{local}$, $Question$)
\ENDIF
\vspace{6pt}
\STATE \textcolor{gray}{\textit{\# Method 2: Ours (Token Confidence)}}

\textbf{LLM\_Generate}($I_{global}$, $Question$)
\IF{Confidence($Logits$) $\ge$ Threshold}
    \STATE \textcolor{blue}{\textit{\# High confidence: Answer directly}}
    \STATE \textbf{Return} $Answer$ 
\ELSE
    \STATE \textcolor{blue}{\textit{\# Low confidence: Extract local details via attention}}
    \STATE $I_{local} \leftarrow$ \textbf{Target\_Guided\_Extract}($I_{global}$)
    \STATE \textbf{Return} \textbf{LLM\_Generate}($I_{global}$, $I_{local}$, $Question$)
\ENDIF
\end{algorithmic}
\end{algorithm}

As shown in Algorithm \ref{alg:workflow}, the LLM-Prompting approach requires a strictly sequential pipeline. The model must explicitly generate a text-based decision before solving the actual task. Consequently, even for simple queries that do not require careful visual inspection, the model wastes computational resources and time performing an additional \textbf{LLM\_Generate} step just to output "No". 

In contrast, our Token Confidence method avoids this bottleneck by attempting to answer the question directly. We calculate the confidence score from the internal probability distribution (Softmax logits) of this first attempt. Since these logits are generated automatically during the forward pass, obtaining the confidence score costs virtually zero extra time. If the confidence is high, we directly output the initial answer, bypassing the need for a second inference round entirely.
\begin{table}[!t]
\centering
\caption{Total inference time comparison between LLM-Prompting and our Token Confidence decision mechanisms (measured on NVIDIA H200 GPUs).}
\footnotesize
\setlength{\tabcolsep}{4.5pt}
\renewcommand{\arraystretch}{1.2}
\resizebox{\columnwidth}{!}{
\begin{tabular}{lccccc}
\toprule
\textbf{Decision Mechanism} & \textbf{AOKVQA} & \textbf{POPE} & \textbf{V*-Bench} & \textbf{HR 4K} & \textbf{HR 8K} \\
\midrule
LLM-Prompting & 16m45s & 1h33m & 1h29m & 1h47m & 3h06m \\
\textbf{Ours (Confidence)} & \textbf{12m33s} & \textbf{1h08m} & \textbf{1h22m} & \textbf{1h36m} & \textbf{3h01m} \\
\bottomrule
\end{tabular}
}
\label{tab:confidence_vs_prompting}
\end{table}
To empirically validate this workflow advantage, we compare the total inference time of the two decision mechanisms. As shown in \autoref{tab:confidence_vs_prompting}, our strategy is consistently faster across all benchmarks under identical hardware conditions. The efficiency gains are particularly pronounced on general reasoning datasets (e.g., saving 25 minutes on POPE, a 26.8\% reduction), where a large portion of queries can be answered using only the global view. Even on challenging datasets like HR-Bench 8K where careful visual inspection is frequently triggered, our method still reduces overall latency by eliminating the redundant "Yes/No" token generation step.

\subsection{Ablation study}
\subsubsection{Ablation on Confidence-based Decision module}
\providecommand{\improvement}[1]{\textbf{\textcolor{red}{+#1}}}
As shown in \autoref{tab:cropping_aokvqa}, our confidence-based decision module consistently improves accuracy compared to indiscriminate cropping (\textit{w/o confidence-based decision}). For Qwen2.5-VL, our strategy yields a +1.14\% on AOKVQA and +0.52\% on V*-Bench. This confirms that indiscriminate cropping introduces visual noise and misjudgments, which our selective strategy effectively avoids.
\begin{table}[H]
\centering
\caption{Quantitative evaluation of confidence-based decision module.}
\footnotesize
\setlength{\tabcolsep}{9pt}
\renewcommand{\arraystretch}{1.1}

\begin{tabular}[!t]{l c c} 
\toprule
\textbf{Model} & \textbf{AOKVQA} & \textbf{V*-Bench} \\
\midrule
Qwen2.5VL-3B \\
\quad \textit{w/o confidence-based decision} & 71.96\% & 85.86\% \\
\quad \textit{w/ confidence-based decision} & \textbf{73.10\%} & \textbf{86.38\%} \\
\rowcolor{gray!20}
\multicolumn{1}{c}{$\Delta$} & \improvement{1.14\%} & \improvement{0.52\%} \\
\bottomrule
\end{tabular}
\label{tab:cropping_aokvqa}
\end{table}
\autoref{tab:cropping_tradeoff} demonstrates the efficiency of our confidence-based decision module. On AOKVQA, we are able to skip redundant cropping for 67.60\% of the samples, effectively halving the inference time without compromising accuracy. Interestingly, this adaptive approach achieves even higher accuracy than applying full cropping (+1.14), indicating that it successfully avoids the interference noise caused by unnecessary local details. For more challenging benchmarks such as V*-Bench, the model naturally performs cropping more frequently, yet it still manages to reduce redundant processing by 22.52\% while sustaining strong performance. These results highlight that LookWise dynamically adjusts its processing behavior according to task complexity, providing a robust and practical trade-off between computational efficiency and predictive accuracy.
\begin{table}[H]
\centering
\caption{Performance and efficiency comparison (measured on NVIDIA H200 GPUs).}
\footnotesize
\setlength{\tabcolsep}{4.5pt}
\renewcommand{\arraystretch}{1.0}

\begin{tabular}[!t]{l c c c}
\toprule
\textbf{Benchmark} & \textbf{Acc} & \textbf{Time} & \textbf{Crop\%} \\
\midrule
AOKVQA\\
\quad \textit{w/o confidence-based decision} & 71.96 & 26m24s & 100\% \\
\quad \textit{w/ confidence-based decision} & \textbf{73.10} & \textbf{12m33s} & 32.40\% \\
\rowcolor{gray!20}
\multicolumn{1}{c}{$\Delta$} 
& \textbf{+1.14} 
& \textbf{-13m51s} 
& \textbf{-67.60\%} \\
\midrule
V*-Bench\\
\quad \textit{w/o confidence-based decision} & 85.86 & 1h04m & 100\% \\
\quad \textit{w/ confidence-based decision} & \textbf{86.38} & 1h18m & 77.48\% \\
\rowcolor{gray!20}
\multicolumn{1}{c}{$\Delta$} 
& \textbf{+0.52} 
& +14m 
& \textbf{-22.52\%} \\
\bottomrule
\end{tabular}
\label{tab:cropping_tradeoff}
\end{table}

\subsubsection{Ablation on Semantic-Guided Localization}
To validate the effectiveness of the semantic-guided localization module, we compare model performance with and without this component (\autoref{tab:mllmsknow_vs_ours}). The quantitative results reveal a clear distinction depending on the task requirements.

On standard visual reasoning datasets (AOKVQA and POPE), the performance improvements are marginal. This is reasonable, as these benchmarks primarily emphasize complex logical reasoning and commonsense knowledge rather than the perception of fine-grained visual details. For such tasks, the bottleneck typically lies in the reasoning capabilities of the language model rather than visual clarity. Since the base global encoding already captures the necessary semantic elements to support this reasoning, extracting high-resolution local crops provides limited additional benefit.

However, on high-resolution and detail-oriented benchmarks (V*-Bench, HR 4K, and HR 8K), the module provides substantial gains. When processing complex scenes, base models often fail to capture small target objects. By incorporating our module, LLaVA-v1.5 improves by 6.78\% on V*-Bench and 4.75\% on HR 8K. The impact is even more pronounced for Qwen2.5-VL, which achieves a 9.96\% boost on V*-Bench and a 6.14\% increase on HR 4K. 

These consistent improvements across different model architectures demonstrate the core advantage of the proposed module. By explicitly using the queried target to guide visual attention, the module isolates task-relevant regions more accurately. This mechanism prevents background noise from interfering with the reasoning process and is crucial for fine-grained localization in complex, high-resolution scenes.
\begin{table}[H]
\centering
\caption{Ablation on semantic-guided localization module.}
\footnotesize
\setlength{\tabcolsep}{6pt}
\renewcommand{\arraystretch}{1.1}

\resizebox{\columnwidth}{!}{
\begin{tabular}{l c c c c c}
\toprule
\textbf{Model \& Strategy} & \textbf{AOKVQA} & \textbf{POPE} & \textbf{V*-Bench} & \textbf{HR 4K} & \textbf{HR 8K} \\
\midrule
\textbf{LLaVA-v1.5-7B} \\
\quad \textit{w/o Semantic-guided Localization} & 72.31\% & 87.25\% & 56.02\% & 44.38\% & 37.25\% \\
\quad \textit{w/ Semantic-guided Localization} & \textbf{72.90\%} & \textbf{87.37\%} & \textbf{62.80\%} & \textbf{47.38\%} & \textbf{42.00\%} \\
\rowcolor{gray!20}
\multicolumn{1}{c}{$\Delta$} 
& 0.59\% 
& 0.12\% 
& \textbf{6.78\%} 
& \textbf{3.00\%} 
& \textbf{4.75\%} \\
\midrule
\textbf{Qwen2.5VL-3B} \\
\quad \textit{w/o Semantic-guided Localization} & 71.62\% & \textbf{89.12\%} & 75.90\% & 66.36\% & 64.88\% \\
\quad \textit{w/ Semantic-guided Localization} & \textbf{71.96\%} & \textbf{89.12\%} & \textbf{85.86\%} & \textbf{72.50\%} & \textbf{70.00\%} \\
\rowcolor{gray!20}
\multicolumn{1}{c}{$\Delta$} 
& 0.34\% 
& -- 
& \textbf{9.96\%} 
& \textbf{6.14\%} 
& \textbf{5.12\%} \\
\bottomrule
\end{tabular}
}
\label{tab:mllmsknow_vs_ours}
\end{table}
\subsection{Confidence Sensitivity Analysis}
\label{sec:Confidence Sensitivity Analysis}

To evaluate the robustness of our confidence-based decision module, we conduct a sensitivity analysis aligned with the two scenarios introduced in ~\autoref{sec:when_to_fuse}: optimizing the decision threshold via the utility function, and applying direct fixed thresholds for open-domain settings.

\subsubsection{Threshold Optimization via Utility Function}
In our formulation, the decision of whether to extract local evidence relies on the utility function $J_{\lambda}(\tau) = \text{TPR}(\tau) - \lambda \cdot \text{FPR}(\tau)$. The hyperparameter $\lambda$ serves as a cost-weighting factor that balances the expected benefit of local inspection against the potential risk of unnecessary computational redundancy. 
Specifically, larger values of $\lambda$ impose stronger penalties on false-positive decisions, encouraging a more conservative local-inspection policy (i.e., yielding a lower optimal threshold $\tau$). Conversely, smaller values of $\lambda$ favor more aggressive local inspection to ensure the capture of fine-grained visual details, thereby expanding the triggering region by adopting a higher threshold.
\begin{table}[H]
\centering
\caption{Sensitivity Analysis of the confidence-based decision module.}
\small
\setlength{\tabcolsep}{6pt}
\renewcommand{\arraystretch}{1.25}
\resizebox{\linewidth}{!}{
\begin{tabular}{l c c c c c c}
\toprule
\multirow{2}{*}{\textbf{Model}} 
& \textbf{Baseline}
& \textbf{Full Local Inspection}
& \multicolumn{3}{c}{\textbf{Confidence-based Decision ($\lambda$)}}
& \textbf{Fixed Thresh.} \\
\cmidrule(lr){4-6}
& (Original)
& (All Inspect)
& \textbf{1.0} 
& \textbf{1.5} 
& \textbf{2.0} 
& ($\tau=0.92$) \\
\midrule
\textit{Threshold ($\tau$)} & - & - & 0.830 & 0.734 & 0.533 & 0.92 \\
\midrule
Qwen2.5-VL 
& 71.44
& 71.96 
& 72.58
& \textbf{73.10} 
& 72.49
& 72.05\\
\bottomrule
\end{tabular}
}
\label{tab:conf_sensitivity}
\end{table}
When representative validation data is available, $\lambda$ can be tuned to maximize task-specific performance. We analyze this optimization on the AOKVQA benchmark by varying $\lambda$. As shown in \autoref{tab:conf_sensitivity}, the performance peaks at 73.10\% when $\lambda = 1.5$, which derives an optimal threshold of $\tau = 0.734$. Notably, this dynamically optimized approach clearly outperforms both the original baseline (71.44\%) and the rigid ``Full Local Inspection'' strategy (71.96\%). This performance gap indicates that indiscriminate zooming not only wastes computational resources but can also introduce visual noise that degrades reasoning. Ultimately, this confirms that our algorithm successfully finds a favorable trade-off between extracting necessary fine-grained details and preventing redundant local inspection.
\begin{table}[H]
\centering
\vspace{-10pt}
\caption{Sensitivity analysis across multiple benchmarks under varying confidence thresholds $\tau$.}
\footnotesize
\setlength{\tabcolsep}{3.0pt}
\renewcommand{\arraystretch}{1.08}
\begin{tabular}{@{}l c cc |cccc@{}}
\toprule
\multirow{2}{*}{\textbf{Dataset}} 
& \multirow{2}{*}{\textbf{Base}}
& \multicolumn{2}{c}{\small \textit{random}}
& \multicolumn{4}{c}{\small \textit{continuous}} \\
\cmidrule(lr){3-4} \cmidrule(lr){5-8}
&
& \textbf{0.80} & \textbf{0.90}
& \textbf{0.94} & \textbf{0.96} & \textbf{0.98} & \textbf{1.00} \\
\midrule
HR-4K    & 67.50 & 72.75 & 72.50 & 72.75 & 72.50 & 72.50 & 72.50 \\
HR-8K    & 58.88 & 68.25 & 69.88 & \textbf{70.00} & \textbf{70.00} & \textbf{70.00} & \textbf{70.00} \\
V*-Bench       & 75.90 & 81.15 & 84.81 & \textbf{86.38} & \textbf{86.38} & 85.86 & 85.86 \\
AOKVQA         & 71.44 & 72.31 & \textbf{72.40} & 72.23 & 72.23 & 72.14 & 71.97 \\
\bottomrule
\end{tabular}
\label{tab:cross_sensitivity}
\end{table}
\subsubsection{Threshold Robustness in Open-Domain Settings}
In real-world open-domain scenarios, it is often impractical to carefully tune the hyperparameter $\lambda$ due to the lack of validation data. To evaluate whether our framework can still perform reliably without such tuning, we sweep across a range of threshold values $\tau$ (see \autoref{tab:cross_sensitivity}). The results show that our method is not sensitive to the exact choice of the threshold. Even without precise calibration, the model maintains stable performance, and its accuracy consistently remains above the baseline across a wide range of relatively high thresholds (e.g., $\tau \in [0.94, 1.00]$).

\begin{table}[H]
\centering
\caption{Evaluation under a fixed threshold ($\tau=0.96$)}
\footnotesize
\setlength{\tabcolsep}{6pt}
\renewcommand{\arraystretch}{1.15}
\begin{tabular}{l ccc}
\toprule
\textbf{Benchmark} & \textbf{Baseline} & \boldmath{$\tau=0.96$} & \textbf{Extraction Reduced} \\
\midrule
AOKVQA           & 71.44 & 72.23 & 10.39\% \\
V*-Bench         & 75.90 & 86.38 & 14.13\% \\
HR-Bench 8K      & 58.88 & 70.00 & 8.00\%  \\
HR-Bench 4K      & 67.50 & 72.50 & 10.00\% \\
\midrule
\rowcolor{gray!15}
\textbf{Average $\Delta$} & -- & \textbf{+9.35\%} & \textbf{10.63\%} \\
\bottomrule
\end{tabular}
\label{tab:fixed_threshold_robustness}
\end{table}
\subsubsection{Robustness with a fixed Threshold} 
Building on the observed stability, we evaluate a single, conservative fixed threshold (\boldmath{$\tau=0.96$}) across all datasets without any per-task adjustment (\autoref{tab:fixed_threshold_robustness}). Under this strictly unified zero-shot setting, LookWise consistently outperforms the baseline, achieving an average accuracy gain of \textbf{+9.35\%}. Simultaneously, it skips redundant local inspection and reduces processed image regions by an average of \textbf{10.63\%}. This confirms that LookWise can function as a robust plug-and-play module that adapts to diverse tasks using a simple fixed threshold.

\subsection{Discussion \& Generalization}
\label{sec:discussion}
\begin{figure*}[!t]
    \centering
\includegraphics[width=0.95\linewidth]{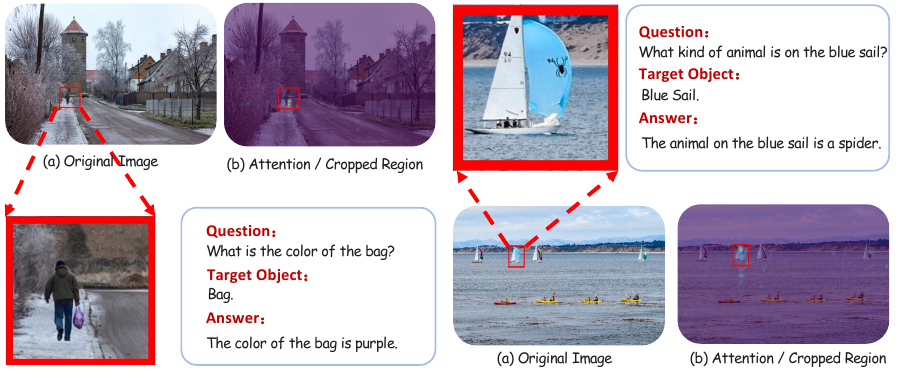}
\caption{\textbf{Qualitative examples of adaptive visual reasoning.} Left: For small object perception, LookWise precisely targets the ``Bag'' to identify its purple color despite a complex background. Right: For fine-grained attribute recognition, the model isolates the ``Blue Sail'' to reveal the subtle spider logo.}
\label{fig:case_bag}
\end{figure*}

\begin{figure*}[!t]
    \centering
    \includegraphics[width=0.90\linewidth]{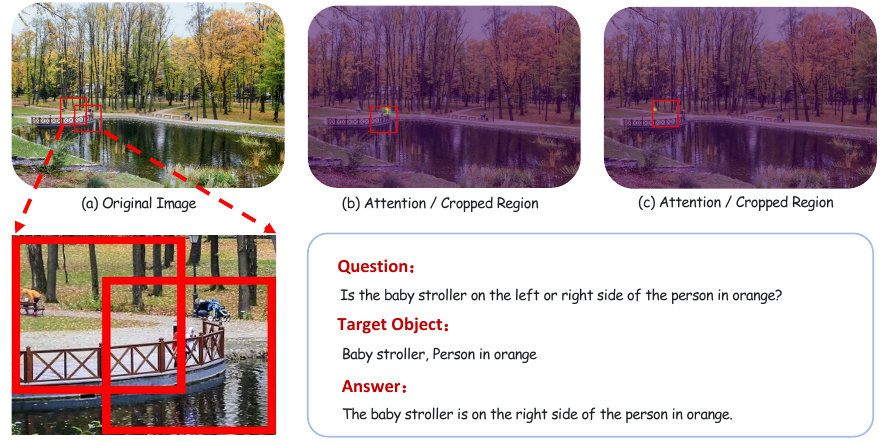}
    \caption{\textbf{Spatial Reasoning with Multiple Targets.} The model attends to both the "Baby stroller" and the "Person in orange," extracting a region that preserves their relative spatial position.}
    \label{fig:case_stroller}
\end{figure*}
\subsubsection{Detailed Performance Analysis on V*-Bench}
While LookWise demonstrates strong performance across most benchmarks, we provide a deeper analysis of its behavior on V*-Bench, focusing on the specific challenges posed by representation constraints and complex spatial reasoning tasks.

\noindent\textbf{Resolution and Representation Constraints.} 
LLaVA-v1.5 processes images at a fixed resolution of $336 \times 336$, which limits the level of visual detail the model can capture. For benchmarks such as V*-Bench that contain many extremely small objects, this resolution becomes a clear limitation. Since each visual token corresponds to a relatively large image region, the resulting attention maps are inherently coarse.
As a result, the model often lacks the precision needed to localize very small targets and produce tight bounding boxes. This limitation is reflected in the much stronger performance observed when using the Qwen2.5-VL. By supporting dynamic high-resolution inputs (up to 3.2M pixels), Qwen2.5-VL provides more detailed visual representations, leading to more precise attention maps and more accurate cropping decisions.

\noindent\textbf{Efficiency vs. Spatial Reasoning Trade-off.} 
V*-Bench heavily features spatial reasoning and visual search queries, such as determining the relative positions of distant objects. While LookWise successfully localizes multiple individual targets, it generates relatively tight bounding boxes around them. When objects are extremely far apart, localizing tightly around them may inadvertently discard the surrounding contextual information necessary for inferring broad relational tasks. 
In contrast, search-based methods like ZoomEye utilize a recursive tree-search: they iteratively split the image into sub-regions, verify targets, and merge confident regions into a single large composite image. This exhaustive approach makes them more robust for complex spatial queries but incurs massive computational overhead (as demonstrated in our efficiency analysis, ZoomEye suffers from an approximately $4.0\times$ latency increase). Overall, LookWise maintains competitive average performance while operating significantly faster. The performance gap on specific V*-Bench spatial queries primarily reflects a deliberate algorithmic trade-off: LookWise prioritizes inference efficiency and precise target-level localization over exhaustive global search.

\subsubsection{Case Studies on Adaptive Visual Reasoning}
\label{sec:casestudy}
We provide qualitative examples (\autoref{fig:case_bag}--\autoref{fig:case_stroller}) to demonstrate how LookWise knows where to look for challenging queries.

\noindent\textbf{Fine-Grained and Small Object Perception.}
Small visual details are easily compressed or lost in standard global low-resolution views. For instance, identifying a minute ``spider'' logo on a sail or a low-contrast bag in a dimly lit street (\autoref{fig:case_bag}) is highly challenging for baseline models. LookWise addresses this by extracting the core textual targets (``blue sail'', ``bag'') and using them to localize subtle query-relevant regions. The resulting high-resolution observations preserve the fine-grained features needed for accurate prediction.

\noindent\textbf{Multi-Target Spatial Reasoning.}
Questions requiring relative position comprehension often cause severe attention drift in standard models, leading them to focus on a single dominant object or irrelevant background. When asking whether a ``baby stroller'' is to the left or right of a ``person in orange'' (\autoref{fig:case_stroller}), LookWise explicitly decouples the two entities. Each textual target produces a corresponding bounding box, and the boxes are merged through a spatial union operation to obtain a local view that covers both subjects. This localized observation preserves their relative spatial relation while filtering out distractors, enabling more reliable spatial reasoning.

\section{Conclusion}
We propose LookWise, a training-free framework for visual reasoning in MLLMs, which can adaptively determines \textit{when} to look more carefully and \textit{where} the query-relevant regions are located.

First, to mitigate the noise caused by blind cropping, we introduce a confidence-based module. In many cases, the global view alone is sufficient for answering simple queries. LookWise therefore estimates uncertainty from the token confidence produced by the initial prediction and triggers cropping only when needed. Second, to address attention drift in current localization methods, we introduce semantic-guided localization. The queried target is extracted from the question and used as the query in a text-to-image attention mechanism, allowing the model to locate relevant regions more accurately.

Extensive experiments demonstrate that LookWise achieves strong performance across multiple challenging benchmarks while maintaining high inference efficiency. As a plug-and-play module that requires no parameter updates, it works across different MLLM backbones. Overall, LookWise reduces perceptual redundancy and alleviates spatial attention drift, highlighting the importance of adaptive visual reasoning for fine-grained multimodal perception.

\bibliographystyle{IEEEtran}
\bibliography{IEEEabrv}

\begin{IEEEbiography}[{\includegraphics[width=1in,height=1.25in,clip,keepaspectratio]{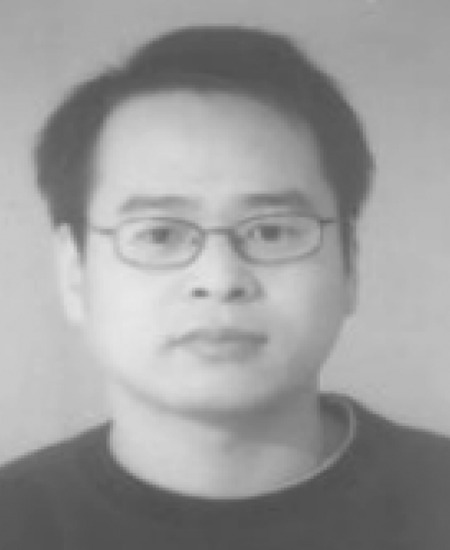}}]{Jie Zhang} received the M.S. degree from the Hefei University of Technology, Hefei, China, in 2009, and the Ph.D. degree from the University of Science and Technology of China, Hefei, in 2014.,He is currently an Associate Professor with the Institute of Intelligent Machines, Hefei Institutes of Physical Science, Chinese Academy of Sciences, Hefei. His current research interests include image processing, pattern recognition, and artificial intelligence.
\end{IEEEbiography}

\begin{IEEEbiography}[{\includegraphics[width=1in,height=1.25in,clip,keepaspectratio]{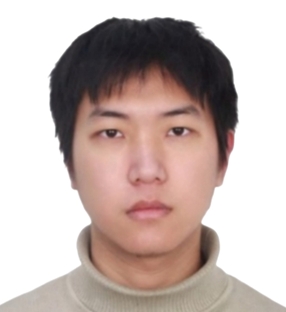}}]{Xuanhua He} received the B.E. degree in software engineering from Xiamen University, Xiamen, China, in 2022. Then He received his M.S. degree at the University of Science and Technology of China, Hefei, China. He is currently pursuing his Ph.D degree with the Hong Kong University of Science and Technology, hongkong, China. His research interests include image restoration and video generation.
\end{IEEEbiography}

\begin{IEEEbiography}[{\includegraphics[width=1in,height=1.25in,clip,keepaspectratio]{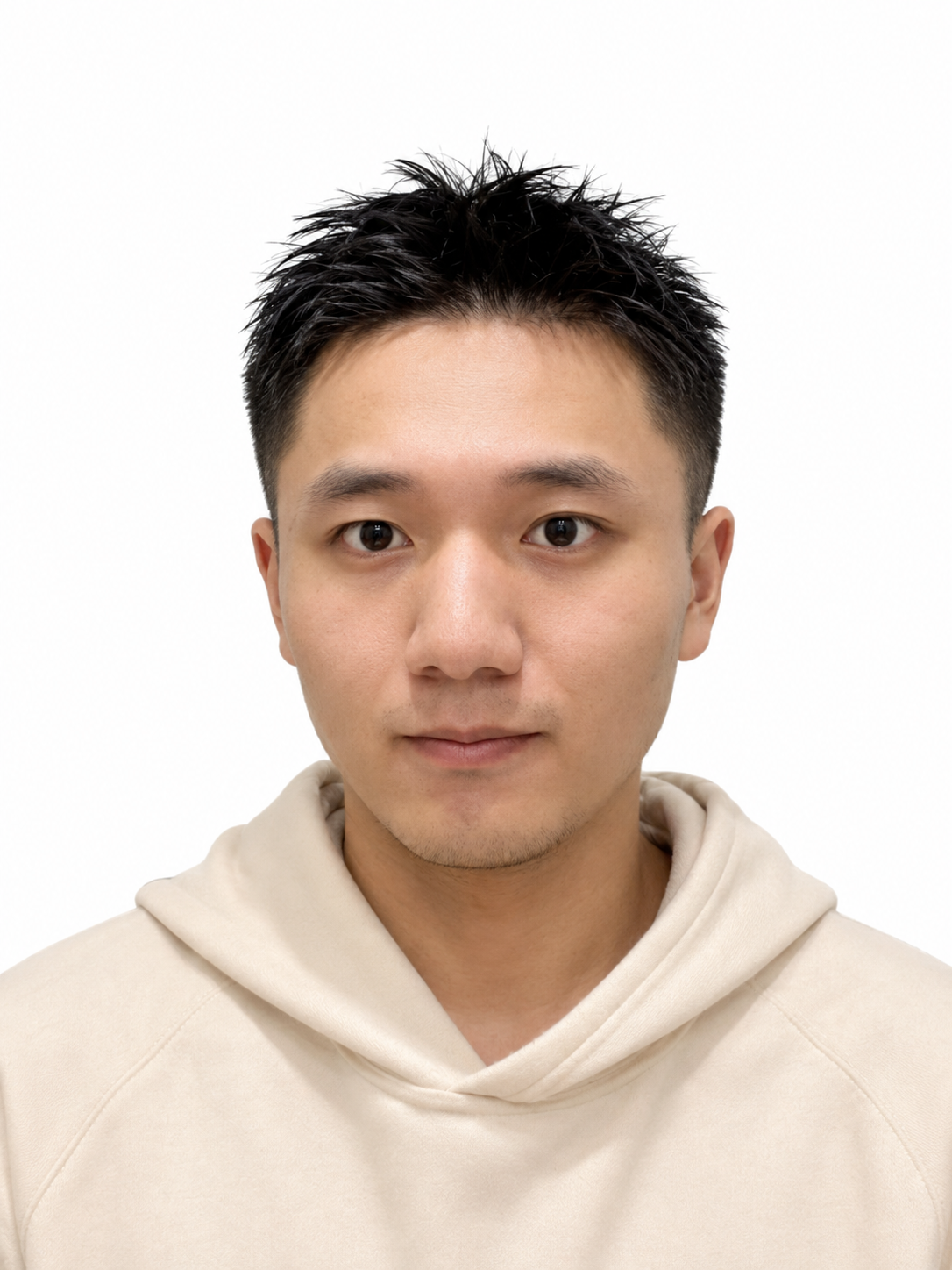}}]{Hailong Huang} received the B.E. degree in computer science and technology from Xiamen University, Xiamen, China, in 2022.Then He received his M.S. degree at the Zhejiang university, Hangzhou, China.
\end{IEEEbiography}

\begin{IEEEbiography}[{\includegraphics[width=1in,height=1.25in,clip,keepaspectratio]{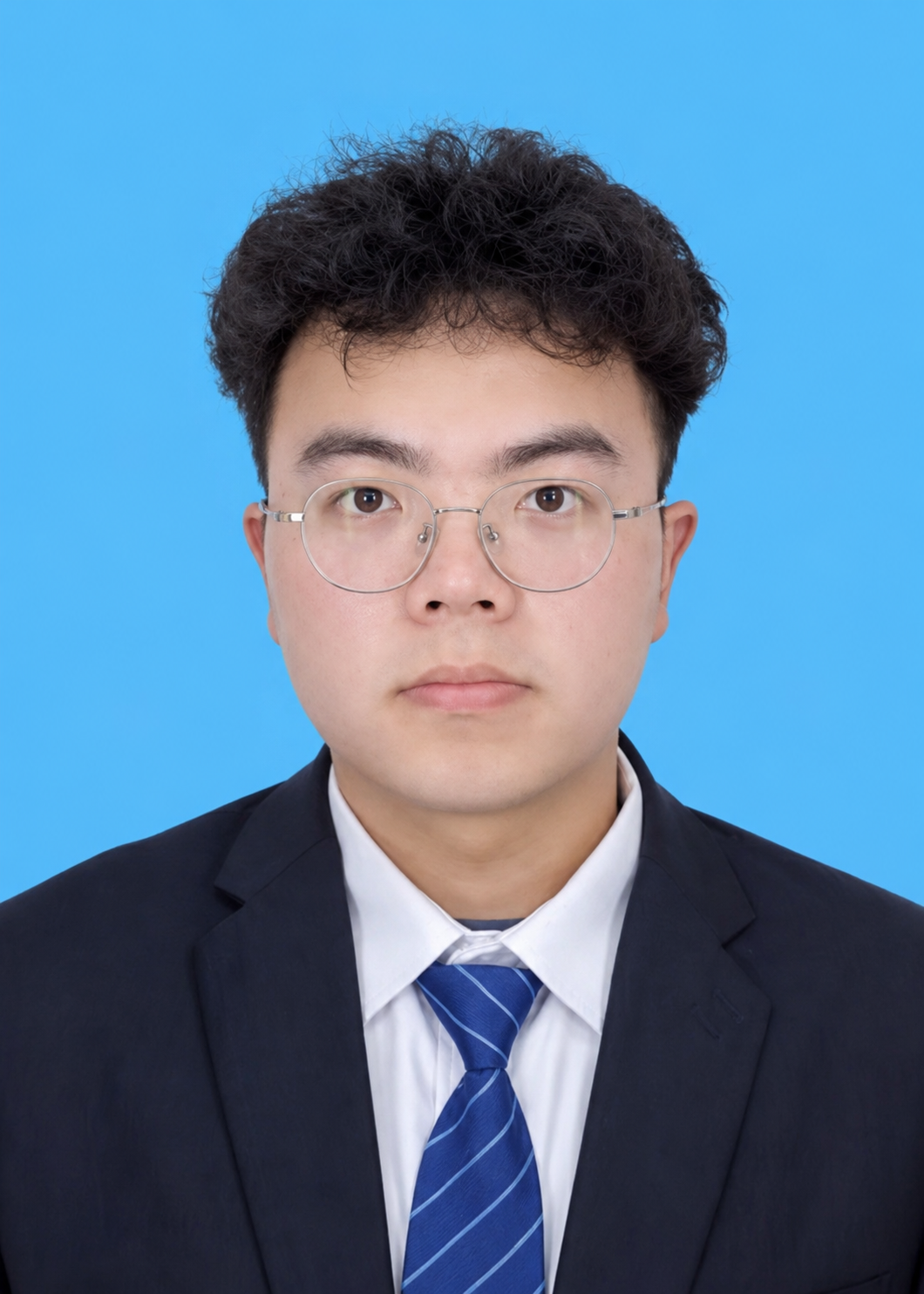}}]{Yuxiang Shen} received the B.E. degree in Electronic Information Engineering from Xiamen University, Xiamen, China, in 2026. He is pursuing the M.S. degree at the University of Science and Technology of China, Hefei, China.
\end{IEEEbiography}

\begin{IEEEbiography}[{\includegraphics[width=1in,height=1.25in,clip,keepaspectratio]{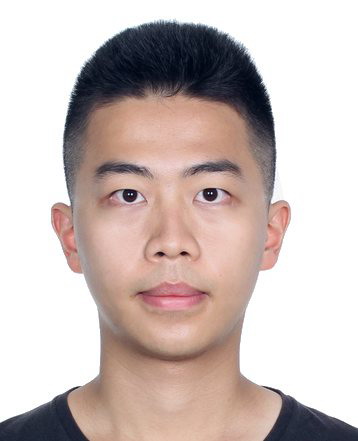}}]{Zhenkun Gao} received the B.E. degree in software engineering from Xiamen University, Xiamen, China, in 2024. He is pursuing the M.S. degree at East China Normal University, Shanghai, China.
\end{IEEEbiography}

\begin{IEEEbiography}[{\includegraphics[width=1in,height=1.25in,clip,keepaspectratio]{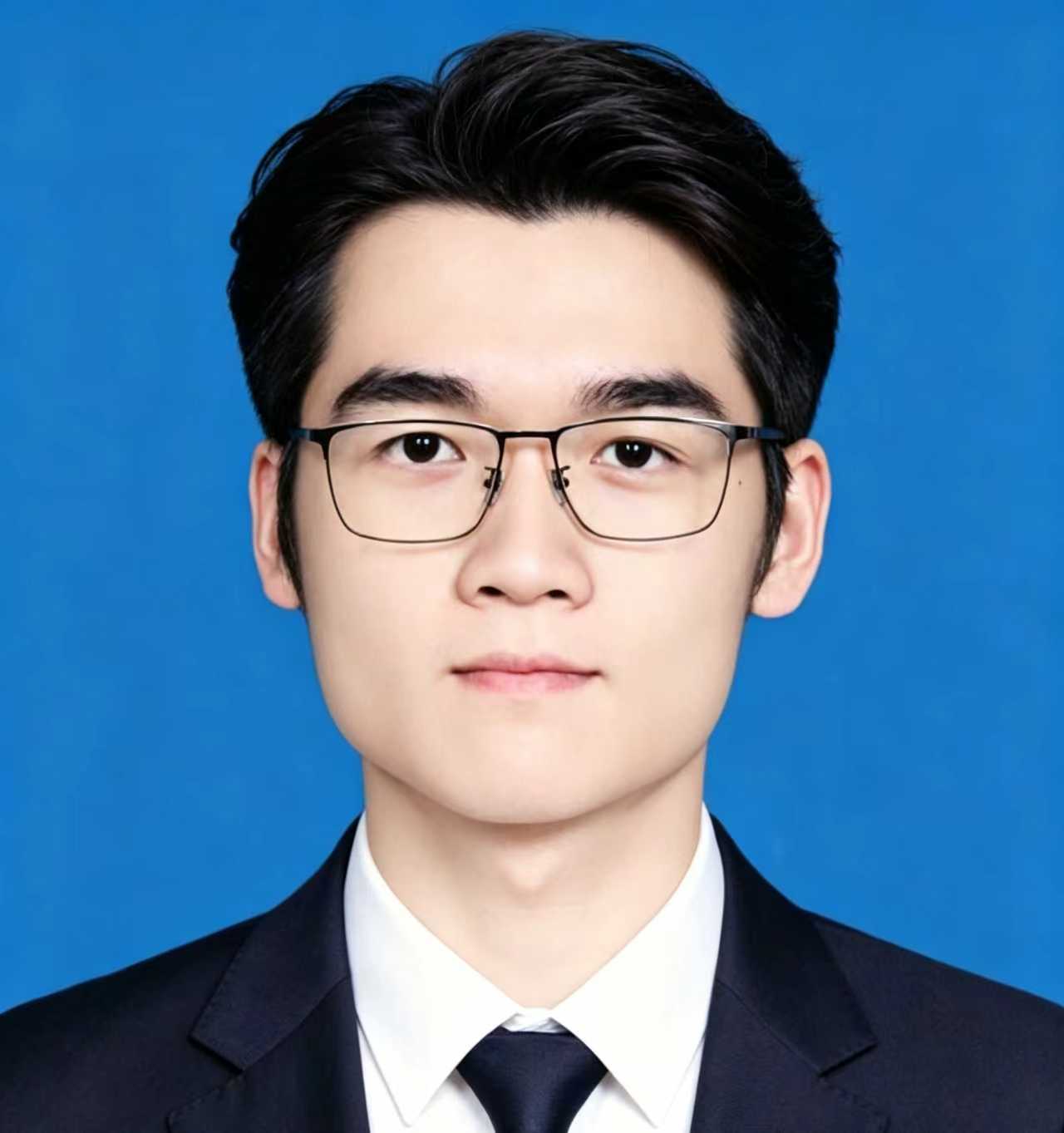}}]{Xueheng Li} received the B.E. degree in computer science and technology from Xiamen University, Xiamen, China, in 2024. He is pursuing the M.S. degree at the University of Science and Technology of China, Hefei, China.
\end{IEEEbiography}

\begin{IEEEbiography}[{\includegraphics[width=1in,height=1.25in,clip,keepaspectratio]{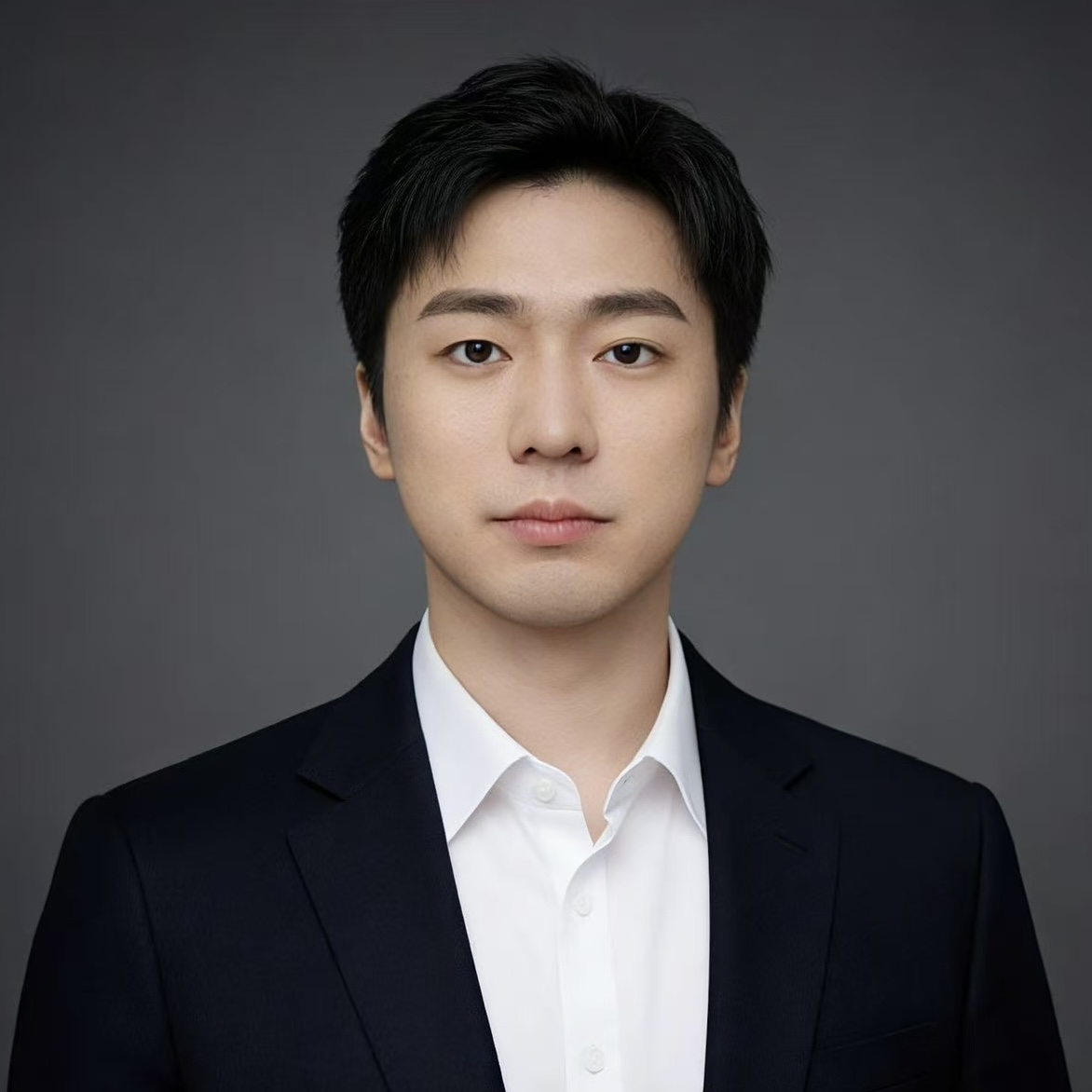}}]{Haoxuan Che} 
received the B.E. degree in Northwestern Polytechnical University (NWPU), Xi'an, China. Then he received Ph.D. in Computer Science and Engineering from The Hong Kong University of Science and Technology (HKUST), Hongkong, China.
\end{IEEEbiography}

\begin{IEEEbiography}[{\includegraphics[width=1in,height=1.25in,clip,keepaspectratio]{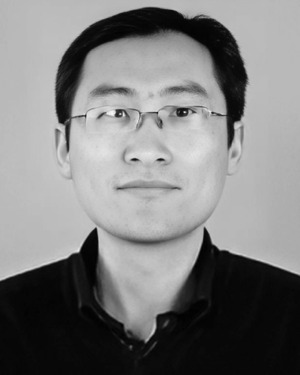}}]{Chengjun Xie} received the M.S. degree in software engineering from the Hefei University of Technology, Hefei, China, in 2008, and the Ph.D. degree from the Hefei University of Technology, Anhui, China, in 2014.,He is working with the Institute of Intelligent Machinery, Chinese Academy of Sciences, Beijing, as an Associate Researcher. His research interests include image processing, machine learning, and pattern recognition.
\end{IEEEbiography}

\begin{IEEEbiography}[{\includegraphics[width=1in,height=1.25in,clip,keepaspectratio]{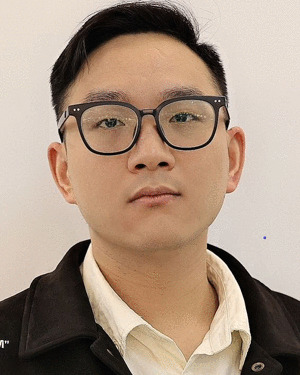}}]{Man Zhou} received the Ph.D. degree from the University of Science and Technology of China, Hefei, China, in 2022. He particularly focuses on geography information system, machine/deep-learning-based satellite image processing, multi-source information fusion. Dr. Zhou was the recipient of the Baidu Scholarship (top ten globally) in 2022 and the WAIC Yunfan Award in 2023 (top 15 globally).
\end{IEEEbiography}
\end{document}